\documentclass{article}

\usepackage[english]{babel}

\usepackage[letterpaper,top=2cm,bottom=2cm,left=3cm,right=3cm,marginparwidth=1.75cm]{geometry}

\usepackage{comment}
\usepackage{subfig}
\usepackage{caption}
\usepackage{tikz}
\usepackage{amsmath, amssymb}
\usepackage{graphicx}
\usepackage{amsthm}
\usepackage{authblk}

\usepackage[colorlinks=true, allcolors=blue]{hyperref}

\newtheorem{definition}{Definition}[section]

\DeclareMathOperator{\SatAgg}{SatAgg}

\title{Neurosymbolic Deep Learning Semantics}
\author[1]{Artur d'Avila Garcez}
\author[2]{Simon Odense}
\affil[1]{City St. George's, University of London}

\begin{document}
\maketitle

\begin{abstract}
\noindent Artificial Intelligence (AI) is a powerful new language of science as evidenced by recent Nobel Prizes in chemistry and physics that recognized contributions to AI applied to those areas. Yet, this new language lacks semantics, which makes AI's scientific discoveries unsatisfactory at best. With the purpose of uncovering new facts but also improving our understanding of the world, AI-based science requires formalization through a framework capable of translating insight into comprehensible scientific knowledge. In this paper, we argue that logic offers an adequate framework. In particular, we use logic in a neurosymbolic framework to offer a much needed semantics for deep learning, the neural network-based technology of current AI. Deep learning and neurosymbolic AI lack a general set of conditions to ensure that desirable properties are satisfied. Instead, there is a plethora of encoding and knowledge extraction approaches designed for particular cases. To rectify this, we introduced a framework for semantic encoding, making explicit the mapping between neural networks and logic, and characterizing the common ingredients of the various existing approaches \cite{SemanticFramework}. In this paper, we describe succinctly and exemplify how logical semantics and neural networks are linked through this framework, we review some of the most prominent approaches and techniques developed for neural encoding and knowledge extraction, provide a formal definition of our framework, and discuss some of the difficulties of identifying a semantic encoding in practice in light of analogous problems in the philosophy of mind. 

\end{abstract}

\section{Introduction}
\subsection{Logic and Neural Networks}
Logic has played a pivotal role in the development of Computer Science and AI. From the beginning of AI research in the 1950's, many believed that the path towards AI lay in the development of computational logical systems capable of emulating cognitive processes. This assumption was built on the computational theory of mind in which cognitive processes are identified with computational ones and logic was an important tool for achieving this \cite{computationaltheory}. This led to the creation of computational logical systems such as logic programming along with related programming languages such as Prolog to develop a framework in which intelligent reasoning could be explicitly defined. However, a competing view of AI was Connectionism.\\\\ Connectionism views the fundamental units of AI systems as being \textit{subsymbolic}, that is, they do not necessarily represent abstract properties that can be reasoned about with simple computational structures \cite{proper_connectionism}. The Artificial Neural Network is the connectionist model put into practice. Neural networks are biologically inspired computational systems consisting of a set of \textit{neurons} which can take certain values, binary ($\{0,1\}$), continuous ($[0,1]$), or otherwise. These neurons are connected to each other in the form of a graph in which the edges between neurons are given \textit{weights} representing the strength of the connection between neurons. At each time step, the state of the network is updated neuron-by-neuron by examining the values of the neighboring neurons, scaling them by their weights, adding them all together, and passing the result through a \textit{transfer function} to produce new values for the neurons, computed in parallel. Although this update is usually applied simultaneously to all neurons in any given network \textit{layer}, it can also be done one at a time. In a network with $n$ neurons, the values of all neurons form a vector in $\mathbb{R}^n$ that we call the \textit{state} of the network. The \textit{state-space} of a neural network is the collection of all possible states. So, if a network has $6$ neurons that each take binary values then its state-space is $\{0,1\}^6$. We use $N$ to denote both a neural network and its update function. That is, given a state, $x$, of network $N$, $N(x)$ refers to the state obtained by using the network to update $x$. The strength of neural networks is in their ability to learn (optimize) the values of the weights from training examples. This is done by subtly adjusting the weights so that the network gets closer to producing some desired behavior. Neural networks can also easily represent training datasets that might be difficult to describe with logic alone, such as natural images. While logic can be used to give an abstract characterization of an image, a neural network can encode it by representing each pixel as a neuron (or 3 neurons for color images). This means that neural networks excel at pattern matching and discovering complicated statistical relationships in the data that might be difficult to express in terms of high-level symbolic variables efficiently.\\\\
However, neural networks have their own shortcomings. The possibly huge number of neurons makes them \textit{opaque} in the sense that it is generally very difficult to understand why a network produced a certain output. They also have trouble generalizing out-of-distribution. That is, dealing with examples that are very different from the ones they were trained on. These are both areas where logic excels: logic is inherently comprehensible by virtue of its compositionality, while its abstract nature makes it generally applicable in a range of situations, familiar or otherwise. For this reason, the field of neurosymbolic AI combines symbolic and subsymbolic systems with the objective of performing both subsymbolic statistical inference and high-level symbolic inference, possibly simultaneously. One avenue towards this is to seek to imbue neural networks with logic. How exactly this is done is something that has been a major research topic in neurosymbolic AI for decades \cite{NeSyBook}. \\\\
On the surface, neural networks and logical systems share little in common. Interestingly, however, the original paper widely credited with establishing artificial neural networks conceived of them as a type of logical system \cite{mcculloch}. Furthermore, recent attempts to encode logical knowledge into neural networks have done so by defining the semantics of logical systems in terms of datasets used to train neural networks \cite{3rdwave}. This is proof that neural networks implement types of logical semantics. In this paper, we explore this concept by discussing the various techniques used to encode logic into neural networks and how they all fit into an overarching framework describing how a neural network can encode the semantics of logic. With this framework in mind, we show how recent work on semantic encoding implicitly define a logical semantics for deep neural networks and, in light of this connection, traditional learning algorithms for neural networks can themselves be thought of as neurosymbolic tasks. We finish by exploring what implications this identification could have on our understanding of deep learning and how viewing deep learning through the lens of logical semantics could potentially provide more information on what makes deep learning successful. \\

\subsection{Neurosymbolic Encoding}
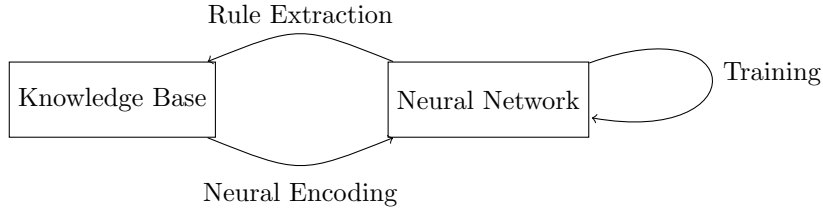
\begin{figure}

\centering
\begin{tikzpicture}

\useasboundingbox (0,0) rectangle (4,4);

 \node[shape=rectangle,draw=black, anchor=center,minimum size=1cm] (y_1) at (-1,2){Knowledge Base} ;

  \node[shape=rectangle,draw=black, anchor=center, minimum size=1cm] (y_2) at (4,2){Neural Network};
  
  \path (y_2) edge [in=350,out=20,loop] node[above, xshift=0.8cm,yshift=-0.2cm]{Training} (y_2);

 \draw[->] (y_2) .. controls (1.5,3) .. (y_1) node[near end, below,yshift = 0.6cm,xshift=0.5cm]{Rule Extraction};
  \draw[->] (y_1) .. controls (1.5,1) .. (y_2)  node[near end, below,yshift = -0.2cm,xshift=-0.5cm]{Neural Encoding};

\end{tikzpicture}
\caption{The neurosymbolic cycle: background knowledge is added to a neural network to help it learn a desired model, rule extraction can then be used to recover the additional symbolic knowledge acquired by the network during training.\label{nesycycle}}
\end{figure}

Much work has been done to find ways to use logic to improve the performance of neural networks. Some of these involve systems with both logical and neural components. Others attempt to create an integrated system in which logical knowledge is contained within a neural network. Henry Kautz in his 2020 AAAI Robert S. Engelmore Memorial Lecture presented a taxonomy of neurosymbolic methods containing 6 categories divided by how closely integrated the symbolic and neural components are. Category 6 includes fully integrated systems in which the logical components are implemented in a neural network and it is this category that we will be discussing in more detail. The idea that logic should, or even could be integrated into a neural network rests on the premise that there is an appropriate logical language to describe the dataset that the network is attempting to learn. We assume that datasets of interest have certain logical regularities that can be exploited by a learning system to generalize beyond the training set. This assumption leads to two fundamental procedures when combining logic and neural networks: encoding and knowledge extraction.\\\\ Encoding of logic into the neural network: if we have prior knowledge of certain logical relationships present in a dataset, then it makes sense to encode these relationships into the neural network before it begins the learning process. This is an intuitive thing to do, try to benefit from existing knowledge, instead of starting from a random network setup. We do not want the neural network to waste time learning relationships in the dataset that we already understand. However, this has to be a flexible process as existing knowledge will frequently require revision, will be partial knowledge potentially containing errors. As pointed out in \cite{bitterlesson}, without such flexibility, results have progressed when networks have been more free to learn patterns on their own without the guidance of preexisting expert knowledge. On the other hand, the limitations of purely data-driven deep learning are becoming hard to ignore \cite{whitepaper}, lack of reliability, data and energy efficiency or fairness, suggesting that there may still be benefit in expert knowledge. As we will see, this neural-symbolic dichotomy may even be a misinterpretation of the underlying processes, as in many important circumstances neural networks innately encode logic of some form in the way that they represent data along with their architecture.\\\\
Extraction of learned knowledge from trained neural networks: this is the inverse problem of the task described above, if we assume that a neural network learns a valid, generalizable interpretation of the data, and we assume that the interpretation has some kind of logical structure, then it is reasonable to assume that the neural network has learned said logical structure. The most common approach to this is known as \textit{rule extraction}, a process that searches for a set of interpretable logical rules that describe accurately the reasoning process of the trained neural network \cite{ruleextraction}.\\\\
Combining encoding and rule extraction leads to the neurosymbolic cycle, depicted in Figure \ref{nesycycle}, in which background knowledge is added to a neural network to help it learn a dataset during training and additional knowledge is extracted from the network after training. The process repeats with the progressive availability of data and knowledge consolidation, with the objective of being parsimonious, promoting modularity, user interaction and the combination of learning and reasoning tools. \\

\begin{figure}
\centering
\begin{tikzpicture}

\useasboundingbox (0,0) rectangle (4,4);

 \node[shape=rectangle,draw=none, anchor=center,minimum width=2.1cm] (x_1) at (-1.5,3.5){Concepts} ;

  \node[shape=rectangle,draw=none, anchor=center, minimum width=2cm] (x_2) at (-1.5,0.5){Knowledge-base};
  
  \node[shape=rectangle,draw=none, anchor=center, minimum width=2cm] (y_1) at (4.5,3.5){Dataset};
  
  \node[shape=rectangle,draw=none, anchor=center, minimum width=2cm] (y_2) at (4.5,0.5){Neural Network};

 \draw[->] (x_1) -- (y_1) node[ below,yshift = 0.5cm,xshift=-3cm]{Satisfied by};
  \draw[<->] (x_2) --  (y_2)  node[ below,yshift = -0.01cm,xshift=-3cm]{Encoding};
  
  \draw[->] (y_1) -- (y_2);
  \draw[->] (x_1) -- (x_2);
  %\draw[->] (x_1) -- (x_2);

\end{tikzpicture}
\caption{Relationship between a neural network trained on a dataset and a knowledge-base representing relations among the concepts of a given task. The question that we wish to investigate is \textit{under what conditions does the encoding of the knowledge base benefit learning with the dataset?}}
\label{hypot}

\end{figure}
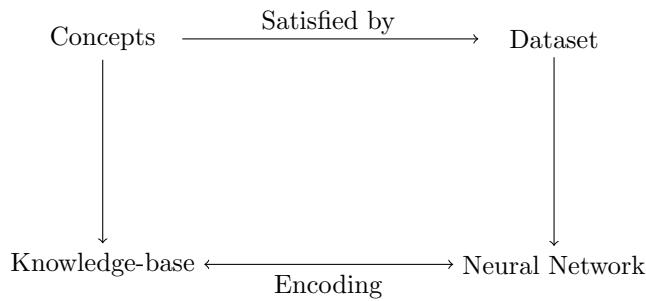

\noindent What should be clear by now is that for the neurosymbolic cycle to have any value, the logic and the knowledge being used should have some clear relationship with the data. This is important in two ways. The first is representational; as we will see in a future section, it is trivial to represent any knowledge base in any logic in a neural network. Without a principled way of representing logic, all neural networks represent all possible logical knowledge. This obviously doesn't get us anywhere. Instead, the way logic is represented must somehow comport with the dataset, in particular what makes the logic represented appropriately with respect to the dataset. The second reason is that the knowledge used should provide utility to the network. If we inject knowledge into a network that completely contradicts the facts of the data, then we are less likely to have a useful neural network at the end of training. Furthermore, even if the knowledge agrees with the data, it should provide information that is not easily gleaned from the data itself; otherwise, the knowledge that we are providing is redundant. In Figure \ref{hypot} we summarize the above requirements in a simple diagram. The conditions we outlined are equivalent to saying that the encoding technique chosen should allow the diagram to commute in the sense that the knowledge base appropriately represents concepts in the dataset and the encoding method allows the network to utilize those concepts when learning the dataset.\\

\noindent Unfortunately, we lack a general set of conditions that ensure the above requirements are satisfied. Instead, various encoding and extraction techniques have been proposed that are designed for particular cases. To rectify this, we introduced a framework for semantic encoding that makes explicit what it means for a neural network to encode logic and identifies the common components of an encoding \cite{SemanticFramework}. We will examine how logical semantics and neural networks are linked through this framework by reviewing the various approaches and techniques developed for neural encoding. Then, we will give a formal definition of the framework and discuss some of the difficulties of identifying a suitable semantic encoding using as reference similar problems from the philosophy of mind. Finally, we will examine how deep learning problems can themselves be seen as a type of semantic encoding and discuss some of the implications of this. To begin, we give an overview of the types of semantic encoding found in the literature.

\section{Types of Encoding Techniques}

The approaches to neural encoding have varied over the years. What is true, however, is that most of them take a \textit{semantic} perspective: the semantics of a logic provides the intended meaning of the encoding. Logical statements are expressed in a formal language and \textit{interpretations} of that language can vary. Although there are notable exceptions such as \cite{TPRorig}, most attempts at encoding logic into neural networks do so by associating particular interpretations with a neural network, specifically its state space. Despite this fundamental underlying similarity, there are several apparently disparate ways of accomplishing this. Initially, strong encoding, which directly encode knowledge into the neural network by setting specific weights, were developed. More recently, neurosymbolic methods have moved towards soft encoding. These methods compile knowledge into a loss function or circuit which can be used alongside or instead of data to train neural networks. The resulting networks often do not encode the knowledge exactly, especially if there are instances in which the knowledge conflicts with the dataset. The resulting networks are generally expected though to conform to the desired solutions more than networks trained on data alone. Below, we use a simple taxonomy to describe approaches to neural encoding. While this identifies the general approaches used for encoding, it does not describe the components involved. As we will see in a future section, all of these approaches share the same basic components and the relationship between the networks and the semantics of logic remains the same in all. For now we describe the encodings.\\\\
\textbf{Strong Encoding:} Strong encoding works by directly designing a neural network to implement a knowledge base. This involves choosing the network architecture, the connections and transfer functions, as well as specifically choosing weights so that the network represents the meaning of the desired knowledge base. For a long time, this was the most common approach to neurosymbolic AI with a focus on encoding logic programs. While strong encodings come with soundness proofs, that is, you know for sure that the knowledge base is encoded in the neural network, the encoding becomes less certain when you begin training the network and, therefore, less useful over time. It is possible that the knowledge base will simply disappear with more and more training from data. One would hope that if the knowledge base is relevant and useful to the data, the learning process would preserve it, but there is no mathematical guarantee of this and, as discussed previously, the usefulness of this encoding will depend heavily on what is encoded and how.   \\\\
\textbf{Equivalences:} Sometimes, an entire class of neural networks is  equivalent to a logical system. In these cases, every neural network in the class can be expressed exactly as a knowledge base of the logical system and vice-versa. Unlike strong encoding, here the neural networks encode by definition a knowledge base of the logic no matter what the weights are. With equivalences, the difference between a neural network and a knowledge base is only representational. The potential benefit is that equivalence to a class of neural networks gives access to efficient learning functions as an implementation of the knowledge bases. How useful this is depends largely on how easily the mapping between network and knowledge base can be identified in practice. A proof that a one-to-one mapping exists between a class of neural networks and knowledge bases does not necessarily tell us how to translate one into the other. Some examples of equivalence include:
\begin{itemize}
        \item Hopfield Networks $\equiv$ Penalty Logic \cite{penalty};
        \item Graph Neural Networks (ACR-GNNs) $\equiv$ $FOC_2$ \cite{GNN:semantic_equiv};
        \item Transformers $\equiv$ First-order Logic with majority classifiers \cite{transformer_equiv}.
\end{itemize} A classic example of equivalence is provided in \cite{penalty} between Hopfield networks and penalty logic. The equivalence comes from the fact that Hopfield networks can be characterized by an \textit{energy function} with the stable states of the network being local minima of the energy function. Penalty logic is a non-monotonic logic developed to handle the possibility of conflicting information. In classical logic, if a knowledge base contains contradictory sentences then the knowledge base will simply map to \textit{false}. In penalty logic, each sentence is given a weight representing the degree of evidence or belief in that sentence. If an interpretation violates a sentence then it incurs a penalty equal to the weight of the sentence. The models of the knowledge base are those with the minimum penalty. It turns out that each knowledge base in penalty logic can be expressed as the energy function of a Hopfield Network and each energy function can be expressed as a penalty logic knowledge base. Other types of equivalences involve the neural network implementing a logical function in each layer, that is, each layer of the network performs the same computation as a function expressed as a logical statement. The network's output is \textit{true} if the logical statement is satisfied, and it is \textit{false} otherwise. This simple idea is adopted to obtain the above equivalence proofs for transformers \cite{transformer_equiv} and graph networks \cite{GNN:semantic_equiv}. \\\\
\textbf{Soft Encoding:} Unlike strong encoding, a soft encoding inserts logical knowledge into the network using one of the main strengths of neural networks: learning. In a soft encoding, logical knowledge is represented either as a loss function or as a circuit \textit{on-top} of the neural network. When the neural network produces an output, the loss on the higher-layers measures how well that output satisfies the desired knowledge base. This information is used to update the weights of the neural network so that its output satisfies the knowledge base. This avoids the problem of strong encoding of knowledge possibly vanishing over time as the knowledge is encoded through learning itself. Furthermore, when data and background knowledge contradict, a soft encoding can find a reasonable compromise between the two. These techniques generally avoid the representational difficulties of semantic encoding by making the language of the logic the same as the language of the dataset. However, the practical problem of choosing good background knowledge for the task at hand persists. This is best exemplified by \textit{reasoning shortcuts}, where a network may find a simple interpretation that technically satisfies a knowledge base even though it is not the desired interpretation \cite{marconato2023neurosymbolic}. \\\\
Something that makes soft encodings even more attractive is that traditional deep learning itself can be seen as a soft encoding. Because soft encodings generally define the semantics of the logic to be encoded in terms of the dataset, the dataset itself can be described logically. We will discuss this in more detail in a subsequent section.
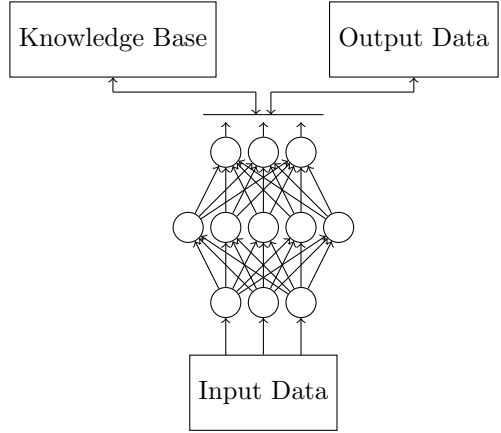
\begin{figure}[htp]

\centering
\begin{tikzpicture}

\useasboundingbox (0,0) rectangle (4,6);
 \node[shape=rectangle,draw=black, anchor=center,minimum size=1cm] (y_2) at (1.5,0.3){Input Data} ;

 \coordinate (y_2l) at (1,0.8);
 \coordinate (y_2r) at (2,0.8);

 \node[shape=rectangle,draw=black, anchor=center,minimum size=1cm] (y_1) at (-0.5,5){Knowledge Base} ;
  \coordinate (kb_1) at (-0.5,4.3);
  \coordinate (kb_2) at (1.4,4.3);
  \coordinate (kb_3) at (1.4,4);
  \draw[<-] (y_1) -- (kb_1);
  \draw[-] (kb_1) -- (kb_2);
  \draw[->] (kb_2) -- (kb_3);

  \node[shape=rectangle,draw=black, anchor=center,minimum size=1cm] (y_3) at (3.5,5){Output Data} ;
\coordinate (od_1) at (3.5,4.3);
\coordinate (od_2) at (1.6,4.3);
\coordinate (od_3) at (1.6,4);

\coordinate (od_4) at (2.3,4);
\coordinate (od_5) at (1.5,4);
\coordinate (od_6) at (0.7,4);

\draw[-] (od_4) -- (od_5);
\draw[-] (od_6) -- (od_5);
\draw[->] (od_1) -- (y_3);

\draw[-] (od_1) -- (od_2);
\draw[->] (od_2) -- (od_3); 
 \node[shape=circle,draw=black, anchor=center,minimum size=0.4cm] (a) at (1,1.5){};
  \node[shape=circle,draw=black, anchor=center,minimum size=0.4cm] (b) at (1.5,1.5){};
   \node[shape=circle,draw=black, anchor=center,minimum size=0.4cm] (c) at (2,1.5){};

  \node[shape=circle,draw=black, anchor=center,minimum size=0.4cm] (i) at (0.5,2.5){};
    \node[shape=circle,draw=black, anchor=center,minimum size=0.4cm] (j) at (1,2.5){};
      \node[shape=circle,draw=black, anchor=center,minimum size=0.4cm] (k) at (1.5,2.5){};
        \node[shape=circle,draw=black, anchor=center,minimum size=0.4cm] (l) at (2,2.5){};
        \node[shape=circle,draw=black, anchor=center,minimum size=0.4cm] (m) at (2.5,2.5){};
  \node[shape=circle,draw=black, anchor=center, minimum size=0.4cm] (d) at (1,3.5){} ;
   \node[shape=circle,draw=black, anchor=center, minimum size=0.4cm] (e) at (1.5,3.5){} ;
 \node[shape=circle,draw=black, anchor=center, minimum size=0.4cm] (f) at (2,3.5){} ;

\draw[->] (y_2) -- (b);
\draw[->] (y_2l) -- (a);
\draw[->] (y_2r) --(c);
 
\draw[->] (a) -- (i);
\draw[->] (a) -- (j);
\draw[->] (a) -- (k);
\draw[->] (a) -- (l);
\draw[->] (a) -- (m);

\draw[->] (b) -- (i);
\draw[->] (b) -- (j);
\draw[->] (b) -- (k);
\draw[->] (b) -- (l);
\draw[->] (b) -- (m);

\draw[->] (c) -- (i);
\draw[->] (c) -- (j);
\draw[->] (c) -- (k);
\draw[->] (c) -- (l);
\draw[->] (c) -- (m);

\draw[->] (i) -- (d);
\draw[->] (i) -- (e);
\draw[->] (i) -- (f);

\draw[->] (j) -- (d);
\draw[->] (j) -- (e);
\draw[->] (j) -- (f);

\draw[->] (k) -- (d);
\draw[->] (k) -- (e);
\draw[->] (k) -- (f);

\draw[->] (l) -- (d);
\draw[->] (l) -- (e);
\draw[->] (l) -- (f);

\draw[->] (m) -- (d);
\draw[->] (m) -- (e);
\draw[->] (m) -- (f);

\coordinate (no1) at (1,3.9);
\coordinate (no2) at (1.5,3.9);
\coordinate (no3) at (2,3.9);

\draw[->] (d) -- (no1);
\draw[->] (e) -- (no2);
\draw[->] (f) -- (no3);

\end{tikzpicture}
\caption{The general structure of a soft encoding. The network is provided with input and output data along with background knowledge. The output of the network is compared to both the desired output data and the knowledge base to check that it is consistent with both. Errors are then used to update the weights of the network so that the output conforms to the desired data and the constraints of the knowledge base.}
\label{nnfig:soft_encoding}
\end{figure}
\\\\
\noindent \textbf{Hard Encodings:} This category could be thought of as a special case of equivalences. The difference here is that, sometimes, a set of neural networks all encode the same knowledge base by virtue of their architecture. One can In this way, research into neural networks has, in certain cases, inadvertently created semantic encodings. A straightforward example of this is the use of the softmax function in feed-forward neural networks. Often neural networks are required to learn how to categorize input examples into one of several categories. This is usually done with a one-hot encoding in which there are $N$ output neurons to represent $N$ distinct categories. The problem is that the transfer functions used by the hidden layers of neural networks do not guarantee that at most one of these neurons will be activated. To rectify this by imposing the required constraint, a softmax layer is used in which the neuron with the highest activation value is selected as the output. If our labels for the output neurons are $Y_1,Y_2,...,Y_N$, we can describe the constraint logically with the following knowledge base:
\begin{displaymath}
\begin{split}
    L=\{Y_1\leftrightarrow \neg Y_2 \wedge\neg Y_3 \wedge ... \wedge \neg Y_N\\
    Y_2\leftrightarrow \neg Y_1 \wedge \neg Y_3\wedge ... \wedge \neg Y_N\\
    \vdots\\
    Y_N\leftrightarrow \neg Y_1\wedge \neg Y_2 \wedge ... \wedge \neg Y_{N-1}\}
\end{split}    
\end{displaymath}
So we see that every neural network with a softmax layer encodes a specific knowledge base. This is further investigated with important results for neurosymbolic AI in \cite{hard_contraint}. Another example of neural architecture implicitly encoding a knowledge base are Convolutional Neural Networks (CNNs). CNNs are designed to capture certain regularities found in natural images. In particular, the layers are designed to be \textit{translation invariant}: no matter how objects in an image are shifted within that image, the network should produce the same output. In practice, most CNNs are not perfectly translation invariant for a number of reasons \cite{CNN_Translation_Invariance}, however, this is another case where the architecture is designed in such a way as to capture a certain property that can be expressed logically. One can express translation invariance as a knowledge base in first-order logic. We will review the semantics of first-order logic in a subsequent section; for now, we simply define a function $f_{shift}$ on images which detects certain objects and shifts them to the left by a fixed amount. If shifting the object results in that object moving outside the boundaries of the image, then the object is wrapped around to the other side of the image. Explicitly defining or coding this function would be extremely difficult. Nevertheless, it exists as a concept which can be defined as follows in first-order logic: 
    \begin{center}
      $f_{shift}(\vcenter{\hbox{\includegraphics[scale=0.02]{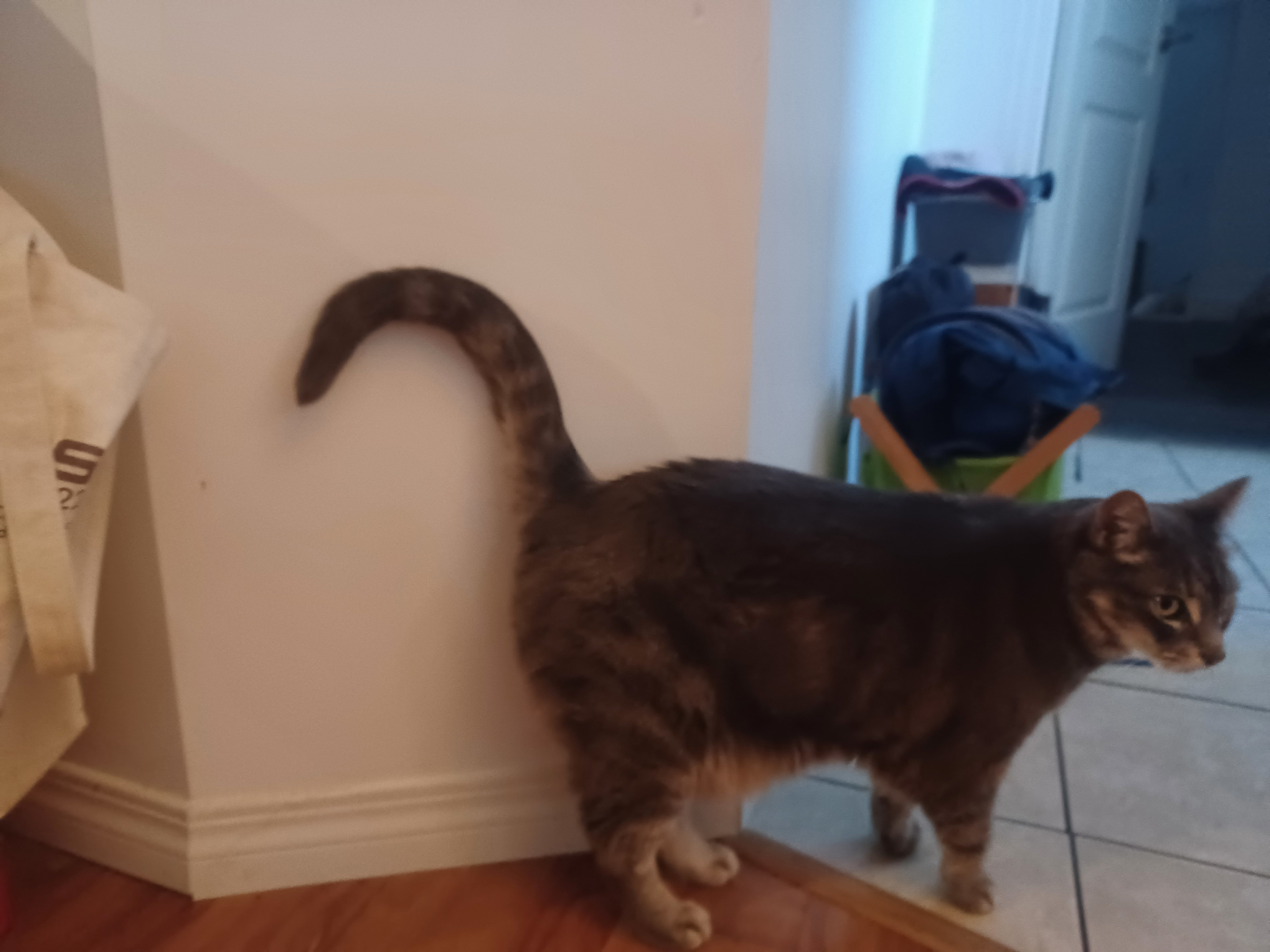}}})=\vcenter{\hbox{\includegraphics[scale=0.02]{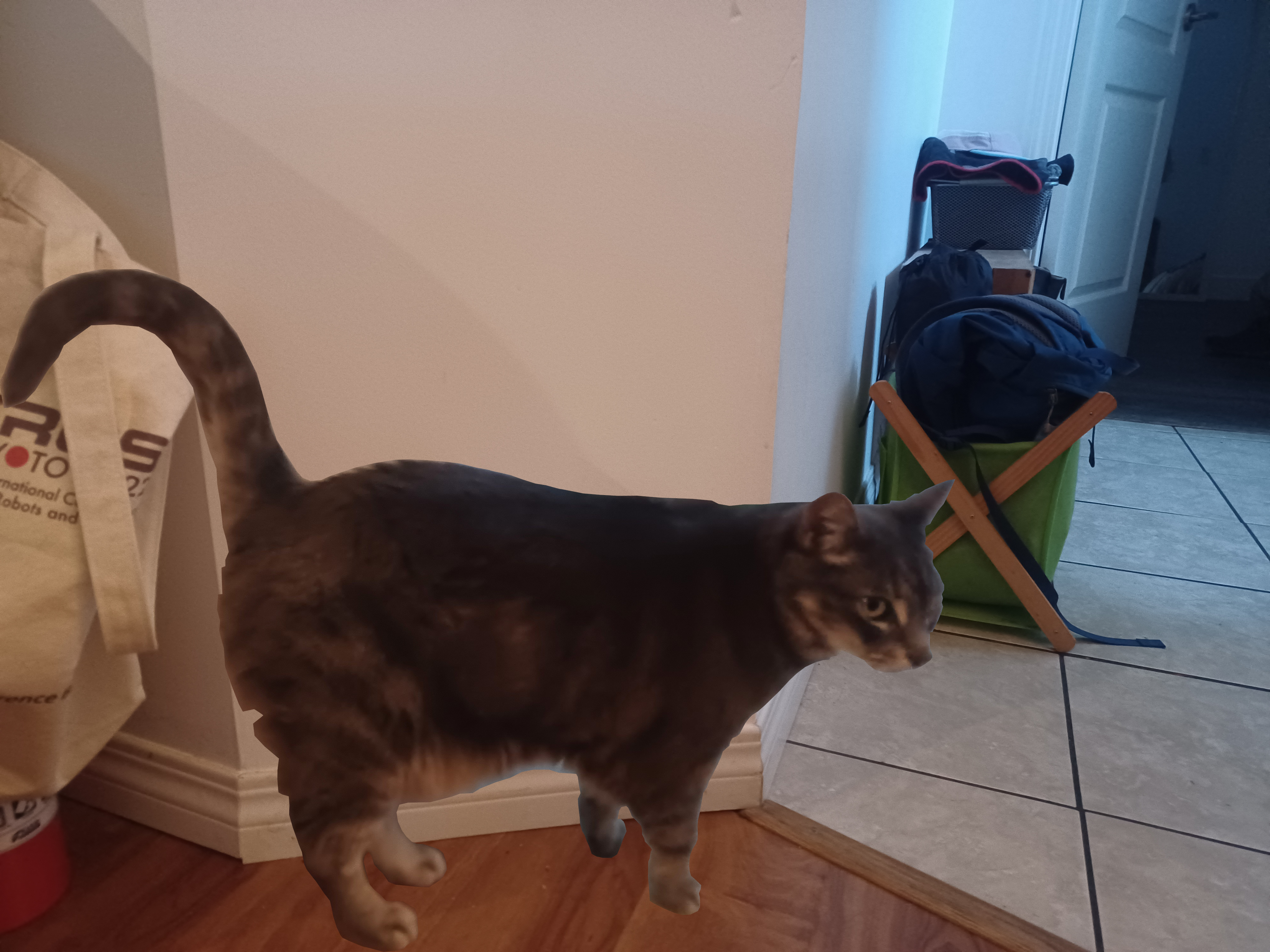}}}$  
    \end{center}
Suppose we wish to detect whether or not a cat is in the picture. We express this as a predicate in first-order logic, $cat(x)$, where $x$ is an image and $cat(x)$ is \textit{true} if $x$ contains a cat. Then, for natural images the following knowledge base is \textit{true}:
\begin{displaymath}
    L=\{\forall x: cat(x)\rightarrow cat(f_{shift}(x))\}
\end{displaymath}
Normally, CNNs will come closer to satisfying this knowledge base than feed-forward networks. We shall give definitions for partially satisfying a logical sentence or knowledge base later. If trained to perfection, CNNs will exactly satisfy the knowledge base. In both cases, the point is that knowledge bases can be used to describe functional or architectural properties of neural network, even if the neural network was not explicitly designed to capture the semantics of any particular logic.\\\\
Now that we know the various approaches of encoding logic into neural networks, we will take a look at how they work in more detail, starting with the classic strong encoding of logic programs. We will briefly review the semantics of propositional and first-order logic before showing how these are used to create strong and soft encodings of knowledge bases. In the following section we will present a generic framework for this process that captures the vast majority of semantic approaches in neurosymbolic AI.
\subsection{Propositional Logic and Logic Programming} 
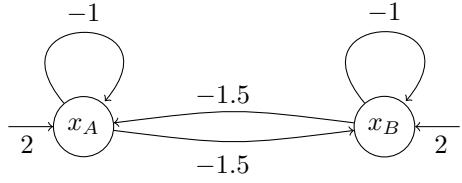
\begin{figure}

\centering
\begin{tikzpicture}

\useasboundingbox (0,0) rectangle (4,4);

 \node[shape=circle,draw=black, anchor=center,minimum size=0.8cm] (y_1) at (0,2){$x_A$};

 \path (y_1) edge [in=50,out=130,loop] node[above]{$-1$} (y_1) ;
 
  \node[shape=circle,draw=black, anchor=center, minimum size=0.8cm] (y_2) at (4,2){$x_B$};
  \path (y_2) edge [in=50,out=130,loop] node[above]{$-1$} (y_2);
 
 \draw[->] (y_2) .. controls (2,2.25) .. (y_1) node[near end, below,yshift = 0.5cm,xshift=0.5cm]{$-1.5$};
  \draw[->] (y_1) .. controls (2,1.75) .. (y_2)  node[near end, below,yshift = -0.1cm,xshift=-0.8cm]{$-1.5$};
  
  \coordinate (b_1) at (-1,2);
    \coordinate (b_2) at (5,2) ;
  
 \draw[->] (b_1) -- (y_1) node[near start, below, xshift=0.1cm]{$2$}; 
 \draw[->] (b_2) --  (y_2) node[near start, below,xshift=-0.1cm]{$2$};
\end{tikzpicture}
\caption{Recurrent neural network with neurons representing propositional atoms $A$ and $B$.}
\label{fig:recurrent_net}
\end{figure}

\begin{figure}
    \centering
    \includegraphics[width=0.9\linewidth]{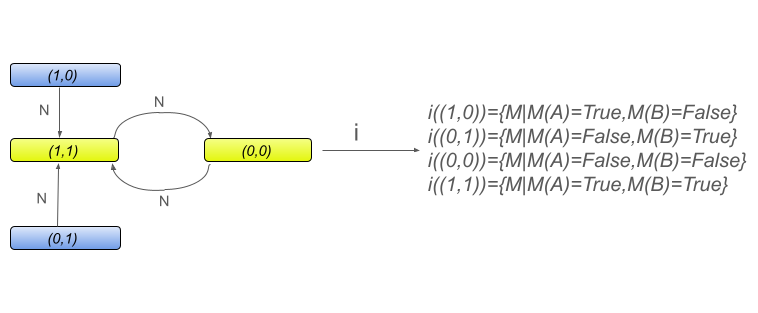}
    \caption{The state transition diagram for the network depicted in figure \ref{fig:recurrent_net} along with the interpretations represented by each state}
    \label{fig:state_transition}
\end{figure}

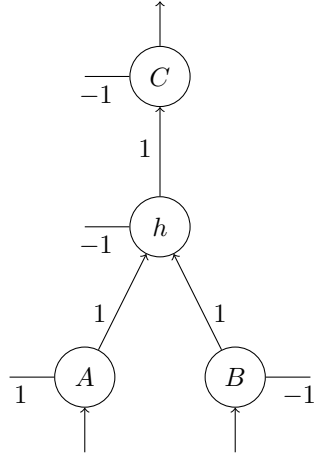
\begin{figure}[htp]

\centering
\begin{tikzpicture}

\useasboundingbox (0,0) rectangle (4,6);

 \node[shape=circle,draw=black, anchor=center,minimum size=0.8cm] (a) at (1,1){$A$};
  \node[shape=circle,draw=black, anchor=center,minimum size=0.8cm] (b) at (3,1){$B$};
 
  \node[shape=circle,draw=black, anchor=center,minimum size=0.8cm] (h) at (2,3){$h$};

  \node[shape=circle,draw=black, anchor=center, minimum size=0.8cm] (c) at (2,5){$C$} ;
  
  \coordinate (b_1) at (0,1);
  \coordinate (b_2) at (1,3);
  \coordinate (b_3) at (1,5);
  \coordinate (b_4) at (4,1);

\coordinate (i_1) at (1,0);
\coordinate (i_2) at (3,0);
\coordinate (o_1) at (2,6);

\draw[->] (a) -- (h) node[ above, xshift=-0.8cm, yshift=-1.39cm]{$1$};

\draw[->] (b) -- (h) node[ above, xshift=0.8cm,yshift=-1.39cm]{$1$};

\draw[->] (h) -- (c) node[ above, xshift=-0.2cm,yshift=-1.2cm]{$1$};

\draw[-] (b_1) -- (a) node[near start, below, xshift=0.0cm]{$1$}; 
\draw[-] (b_2) --  (h) node[near start, below,xshift=-0.0cm]{$-1$};
\draw[-] (b_3) --  (c) node[near start, below,xshift=-0.0cm]{$-1$};
  
\draw[-] (b_4) -- (b) node[near start, below,xshift=-0.0cm]{$-1$};

 \draw[->] (i_1) -- (a) node[near start, below, xshift=0.0cm]{}; 
 \draw[->] (i_2) -- (b) node[near start, below, xshift=0.0cm]{}; 
 \draw[->] (c) -- (o_1) node[near start, below, xshift=0.0cm]{}; 

\end{tikzpicture}
\caption{A simple feed-forward neural network encoding a knowledge-base containing logic programming rules \textit{C if A}, written  $C\leftarrow A$, \textit{C if B}, written $C\leftarrow B$, and a fact \textit{A}, written $A\leftarrow $. The parameters of the network (weights and biases) are shown next to the arrows in the diagram. With bias $1$, neuron \textit{A} will always produce output $1$ (we say that \textit{A} is \emph{activated} in this case) for any input in $\{0,1\}$ given a step function as activation function. With bias $-1$, neuron \textit{B} will output $0$ for every input. Activating either \textit{A} or \textit{B} will always activate hidden neuron \textit{h}, since the weight ($1$) from either \textit{A} or \textit{B} to \textit{h} is equal  to (or larger than) the negative of the bias of \textit{h}. Finally, activating \textit{h} will also activate \textit{C}, for the same reason.}
\label{nnfig:1}
\end{figure}

Most early attempts at combining logic with neural networks focused on encoding various forms of propositional logic directly into a neural network. %In fact, this was a source of criticism for neural networks as it was alleged that they had an inherent``propositional fixation" the individual neurons in a neural network were best understood as propositional atoms meaning that neural networks lacked the type of constitutionality inherent in other types of more complicated logic, notably first-order logic. (%AG: the "fixation" was claimed by John McCarthy as early as 1988; we probably don't need to get into this.)
While in the next section we will cover first-order and many-valued logic, for now let us look at the wide array of techniques developed to encode propositional knowledge in neural networks.\\\\
Propositional logic is concerned with the logical relationships between atomic \textit{propositions}. Each proposition can be assigned a truth value - \textit{true} or \textit{false} - and multiple propositions can be combined to form more complex statements whose own truth value can be fully determined from the truth values of its constituent atoms. For example, a proposition may be something like \textit{Bob likes to eat apples} while another proposition may state that in order to eat an apple Bob has to have an apple, known as a pre-condition in the context of planning in AI. From these we can form the sentence: if \textit{Bob likes to eat apples} (A) and \textit{Bob has an apple} (B) then \textit{Bob is likely to eat an apple} (C). Assuming the truth of A and B, we may expect to deduce C. Expressing this as a single sentence in propositional logic we have $(A \wedge B) \rightarrow C$. We construct the language of propositional logic from the following symbols: a set of atomic propositions known as atoms, $\{A,B,C,...\}$, and a set of logical connectives, $\{\vee,\wedge,\neg,\rightarrow\,\leftrightarrow\}$. A sentence in propositional logic is constructed recursively according to the following rules:
\begin{itemize}
    \item If $A$ is an atom then it is a sentence in propositional logic;
    \item If $\phi$ is a propositional sentence then $\neg \phi$ is a propositional sentence;
    \item If $\phi_1$ and $\phi_2$ are propositional sentences then $\phi_1\wedge \phi_2$, $\phi_1\vee \phi_2$, $\phi_1\rightarrow \phi_2$ and $\phi_1\leftrightarrow \phi_2$ are also propositional sentences.
\end{itemize} 
Here, $\wedge,\vee,\neg,\rightarrow,\leftrightarrow$ are interpreted as logical \textit{and}, \textit{or}, \textit{negation}, \textit{implication} and \textit{if and only if}, respectively. In order to evaluate the truth of a sentence in propositional logic, we must first know the truth of the atoms contained in the sentence. This makes the logical language compositional. We define an interpretation of propositonal logic as a mapping $M:\mathcal{A}_{prop}\rightarrow \{True,False\}$, where $\mathcal{A}_{prop}$ is the set of atoms. Each interpretation extends to map $\hat{M}:\mathcal{L}\rightarrow\{True,False\}$, where $\mathcal{L}_{prop}$ is the set of propositional sentences by recursively applying the following rules:
\begin{itemize}
    \item if $\phi=A$ then $\hat{M}(\phi)=M(A)$;
    \item if $\phi=\neg \phi_0$ then $\hat{M}(\phi)=True$ if $\hat{M}(\phi_0)=False$ and $False$ otherwise;
    \item if $\phi=\phi_1 \wedge \phi_2$ then $\hat{M}(\phi)=True$ if $\hat{M}(\phi_1)=True$ and $\hat{M}(\phi_2)=True$;
    \item if $\phi=\phi_1\vee\phi_2$ then $\hat{M}(\phi)=True$ if $\hat{M}(\phi_1)=True$ or $\hat{M}(\phi_2)=True$;
    \item if $\phi=\phi_1\rightarrow \phi_2$ then $\hat{M}(\phi)=True$ if $\hat{M}(\phi_1)=False$ or if $\hat{M}(\phi_1)=True$ and $\hat{M}(\phi_2)=True$;
    \item if $\phi=\phi_1\leftrightarrow \phi_2$ then $\hat{M}(\phi)=True$ if $\hat{M}(\phi_1\rightarrow\phi_2)=True$ and  $\hat{M}(\phi_2\rightarrow\phi_1)=True$.
\end{itemize}
The sentence $A\rightarrow B$ is read \textit{if} $A$ \textit{then} $B$. If $A$ is false then this implication is trivially \textit{true}. If $A$ is \textit{true}, however, then $B$ must not be \textit{false} for the implication to hold. Extensions of propositional logic allow for more than two truth values. Some systems introduce a third value representing \textit{unknown} while others assign a real number to each atom representing a degree of belief. Some give an arbitrarily finite number of possible truth values. In all cases, the logical connectives must be interpreted in a way that preserves their intended meaning. For now we stick with binary truth values. \\\\
We define a knowledge base in propositional logic, $L$, as a set of propositional sentences. In other words $L\subset 2^{\mathcal{L}_{prop}}$. An interpretation, $\hat{M}$ maps a knowledge base to $True$ if it maps each sentence in the knowledge base to $True$, otherwise it maps the knowledge base to $False$. The \textit{models} of $L$ consist of those interpretations that map it to $True$.\\\\
For neural networks whose neurons take binary values, there is an obvious way to assign interpretations to states of the network by identifying neurons with atoms. If the activation value of a neuron is $1$ then the corresponding atom has truth value $True$ and if the value is $0$ then the corresponding atom has truth value $False$. Take the network in Figure \ref{fig:recurrent_net} for example.
Assume that the values of the neurons are binary with the standard step function for their transfer function. Here, we identify $x_A$ with the atom $A$ and $x_B$ with the atom $B$. The state of the network is given by the value of $(x_A,x_B)$. There are 4 possible states: $(0,0),(0,1),(1,0),(1,1)$. With the neurons-as-atoms identification described above, these should correspond, respectively, to interpretations in which $M(A)=M(B)=False$, $M(A)=False$ and $M(B)=True$, $M(A)=True$ and $M(B)=False$, $M(A)=M(B)=True$. Note that in the future, for simplicity, we will often drop the notation $M$ when referring to the truth value of individual atoms. The statement $A=True$ should be understood as meaning that $M(A)=True$ for a given model $M$.\\\\ One complication here is that in our formulation, interpretations of propositional logic refer to truth-assignments to \textit{all} atoms and not just a subset of them. The network in question has no neurons representing atom $C$ or any of the infinite atomic propositions in the language. If we were dealing with a language with unknown truth values, we may simply assign $Unknown$ to each atom not represented by the network. Since we are not, a simple work around is for each state to represent the set of interpretations which satisfy the condition imposed by the network. In other words, the state $(0,1)$ represents the set $\{M|M(A)=False,M(B)=True\}$. In this way, each state represents a set of possible worlds, the values of $x_A$ and $x_B$ fix the truth values of the atoms $A$ and $B$ but say nothing about the truth values of other atoms. We will use $i$ to denote mappings from network states to sets of interpretations. Figure \ref{fig:state_transition} shows the state transition diagram of the network along with the sets of models corresponding to each state. Notice that no matter the state in which the network starts, it ends in the cycle $(1,1)\rightarrow(0,0)\rightarrow (1,1)$. This suggests that the interpretations represented by the states $(0,0)$ and $(1,1)$ are of special importance. This is something that we will return to when we define the semantics of a neural encoding.
\\\\
While propositional logic is the backbone of much of the logic used in AI, it was logic programming that became the focus of early encoding techniques.
Logic programs consist of sentences of the form $A\leftarrow B_1\wedge B_2\wedge ... \wedge B_n$ where each $B_i$, called the \textit{body} of the logic programming rule, is either an atom or the negation of an atom; $A$ is called the \textit{head} of the rule (the symbol $\leftarrow$ is used to differentiate from the classical implication interpretation seen before). In general, the atoms in logic programming are grounded predicates of first-order logic. However, the neural encoding of logic programs that we wish to discuss generally assume that the domain is finite, meaning that a first-order logic program can be thought of as a short hand for a larger propositional logic program. Interpretations in a logic program are the same as in propositional logic with the main difference that a logic program defines a fixed-point operator over $\leftarrow$ to make $A$ $True$ whenever $B_1, B_2, ..., B_n$ are $True$. Given an interpretation, $M$, and a logic program $P$, the fixed-point operator $T_P:\mathcal{M}\rightarrow \mathcal{M}$ is defined as $T_P(M)=M'$ where $M'(A)=True$ if and only if $P$ contains a sentence $A\leftarrow B_1\wedge B_2\wedge ... \wedge B_n$ with $M(B_i)=True$ for $0\leq i \leq n$. This also allows for rules with empty bodies, $A\leftarrow$, where the condition is trivially satisfied for every $M$, making $A$ a ground fact, that is $True$. The models of a logic program are those interpretations which are fixed under $T_P$. In other words $M$ is a model of $P$ if $T_P(M)=M$. This is operationalized by starting from the facts and applying the rules in the direction of the arrow ($\leftarrow$), adding the heads of the satisfied rules to the set of facts until there are no more facts to add. Whether or not $T_P$ converges to a fixed point depends on the structure of the logic program, in particular, acyclic logic programs will always converge to a fixed point whereas programs that contain cycles may not (logic programs can have their rules sorted hierarchically such that atoms in one level of the hierarchy do not appear in the head of any rule in a higher level of the hierarchy) .\\\\ The focus of neural encodings of logic programs was to define or learn $T_P$ as the function computed by a neural network. The first neurosymbolic learning system to do so was the Connectionist Inductive Logic Programming (CILP) system \cite{CILP}, inspired by the earlier Knowledge-based Neural Networks (KBANN) \cite{KBANN} which looked specifically at acyclic Horn clauses, that is acyclic logic programs with no negation. Like in our previous example, this was done by identifying neurons with individual atoms and using a recurrent neural network in the case of CILP or a feed-forward neural network in the case of KBANN to compute the truth values of the heads of rules from the truth values of their bodies. Figure \ref{nnfig:1} illustrates the general principle.\\

\noindent In the network of Figure \ref{nnfig:1}, the logic program is $P=\{C\leftarrow A,  C\leftarrow B, A \leftarrow\}$. Given any propositional interpretation, say $I=\{A=True,B=False,C=False\}$, we can use this program to update the interpretation, $T_P(I)=I'=\{A=True,B=False,C=True\}$. Iterating again, we see that $T_P(I')=I'$ and thus that $I'$ is the least-fixed point of this program and the unique model of the program. Notice that $I'$ is also a model of the standard propositional knowledge base $\{A\rightarrow C,B\rightarrow C, A\}$. Now consider the equivalent computation done in the network. As a choice, take $A=1,B=0,C=0,h=0$ to be the activation values of the neurons to start with. Updating the network activations once in the usual way will give $A=1,B=0,C=0,h=1$. Updating it again gives $A=1,B=0,C=1,h=1$ which is a stable state. We can see that the network is an implementation of $T_P$ and thus an encoding of $P$.\\\\
The idea of using a neural network to implement the least fixed-point operator became a staple approach to encoding logic programs and the resulting neurosymbolic methods became known as CORE methods \cite{CORE}. Originally, the goal was executing the logic program in parallel using the network for the purpose of efficient theorem proving. Later the method was extended to allow the combination of learning and reasoning with the use of differentiable activation functions when the CILP system combined the CORE method with backpropagation \cite{CILP}. Most advances that followed in this area dealt with encoding different types of logic programs that may have more complicated behavior or be more expressive such as answer set programming, epistemic and temporal logic programs \cite{NeSyBook}. To encode logic programs which may not be acyclic, CILP uses recurrent connections linking up neurons appearing with the same label both in the output and input layers. If the logic program being encoded has a least-fixed point then CILP will converge to it. Many techniques adopted the same CORE approach but allowed for multi-valued logic programs in which truth values are no longer binary but can take on three or more values. CILP's extension to first-order logic programs adopts a multi-valued approach \cite{CILP++}, with other methods of translating least-fixed point operators into multi-layered neural networks including \cite{multivalue}.\\\\ 
All of the above fall under the category of strong encoding. Most soft-encoding techniques focus on first-order logic, although semantic regularization \cite{sem_reg} is a soft encoding technique that uses binary-valued output neurons trained with a loss function representing the probability that the labels satisfy some condition. An example of a condition is the one we have already discussed with the softmax layer. If a neural network has $n$ output neurons, $y_1,y_2,...,y_n$ with a softmax layer, let $Y_i$ denote the atom corresponding to neuron $y_i$, $1 \leq i \leq n$. Each state of the neural network encodes a set of propositional interpretations $\{M|Y_i=True\leftrightarrow y_i=1\}$. The softmax layer ensures that, for all $i$, the sentence $\phi_i =Y_i\leftrightarrow \neg Y_1 \wedge ... \wedge \neg Y_{i-1}\wedge \neg Y_{i+1} \wedge ... \wedge \neg Y_n$ is encoded by the network. Without a softmax layer, this property would not hold in general. With semantic regularization, the output neurons are given values in $[0,1]$ interpreted as the probability of being activated. With these values you can measure, for every possible output value, the probability that $\phi_i$ is satisfied and use this to change the weights of the neural network. By doing this a neural network is able to learn to satisfy, at least approximately, $\phi_i$. We can obviously repeat this for any propositional knowledge base we wish. This is a slightly unusual example of a soft neurons-as-atoms semantic encoding of propositional knowledge bases, because as mentioned most soft encoding techniques focus on first-order logic, especially fuzzy and probabilistic variants. We will discuss these next. 

\subsection{First-Order Logic, Fuzzy and Probabilistic Logic} 

\begin{figure}
\begin{tikzpicture}[scale=0.95]
\centering
\useasboundingbox (0,0) rectangle (10,10);

       \node[anchor=center,minimum size=0.8cm,label={[align=left]$ (R_1(x,y)\wedge \neg R_2(y))\rightarrow R_2(x)$}] (q2) at (8,8){} ;

           \node[anchor=center,minimum size=0.8cm,label={[align=left]$ R_1(x,y)\wedge \neg R_2(y)$}] (q3) at (6,6.5){} ;

               \node[anchor=center,minimum size=0.8cm,label={[align=left]$ R_2(x)$}] (q4) at (11,6.5){} ;
   \node[anchor=center,minimum size=0.8cm,label={[align=left]$ \rightarrow $}] (nolab) at (9,6.3){} ;

    \node[anchor=center,minimum size=0.8cm,label={[align=left]$ R_1(x,y)$}] (q5) at (4,5){} ;

       \node[anchor=center,minimum size=0.8cm,label={[align=left]$ \wedge$}] (nolab) at (6,4.7){} ;

    \node[anchor=center,minimum size=0.8cm,label={[align=left]$\neg R_2(y)$}] (q6) at (8,5){} ;
    \node[anchor=center,minimum size=0.8cm,label={[align=left]$R_2(y)$}] (q7) at (8.1,4){} ;

    \node[anchor=center,minimum size=0.8cm,label={[align=left]$\neg$}] (nolab) at (8.4,4.65){} ;
\coordinate (c_1) at (8,5.1);
\coordinate (c_2) at (8,5.5);
\draw[->] (c_1) -- (c_2);
%$\{M|R((x_{y_1,1},x_{y_1,2}),(x_{y_2,1},x_{y_2,2}))$\\$=x_{Ry_1y_2}\}$

    %$    \node[anchor=center,minimum size=0.8cm] (g) at (7,1){$(1,2,2,2)$} ;

 \coordinate (b_1) at (7.8,7.5);
 \coordinate (b_2) at (10.3,7.5);
 \coordinate (b_3) at (9,7.5);
 \coordinate (b_4) at (9,8.5);

  \draw[-] (b_1) -- (b_3);
  \draw[-] (b_2) -- (b_3);
  \draw[->] (b_3) -- (b_4);

  \coordinate (a_1) at (6,5.8);
  \coordinate (a_2) at (5,5.8);
  \coordinate (a_3) at (7,5.8);
  \coordinate (a_4) at (6,7);
  \draw[-] (a_2) -- (a_1);
  \draw[-] (a_3) -- (a_1);
  \draw[->] (a_1) -- (a_4);

 \node[anchor=center,minimum size=0.5cm] (x) at (6,0.2){$x$} ;

 \node[anchor=center,minimum size=0.5cm] (y) at (10,0.2){$y$} ;

\coordinate (i_1) at (7.5,0.7);
\coordinate (i_2) at (8,0.7);
\coordinate (i_3) at (8.5,0.7);

\draw[-] (i_1) -- (i_2);
\draw[-] (i_3) -- (i_2);

\draw[black] (4.5,0) rectangle (7.5,0.5);

  \draw[black] (8.5,0) rectangle (11.5,0.5);

 \node[shape=circle,draw=black, anchor=center,minimum size=0.05cm] (nf1) at (7.5,1.2){};
  \node[shape=circle,draw=black, anchor=center,minimum size=0.05cm] (nf2) at (8,1.2){};

   \node[shape=circle,draw=black, anchor=center,minimum size=0.05cm] (nf3) at (8.5,1.2){};

  \node[shape=circle,draw=black, anchor=center,minimum size=0.05cm] (ns1) at (7,2.2){};
    \node[shape=circle,draw=black, anchor=center,minimum size=0.05cm] (ns2) at (7.5,2.2){};
      \node[shape=circle,draw=black, anchor=center,minimum size=0.05cm] (ns3) at (8,2.2){};
        \node[shape=circle,draw=black, anchor=center,minimum size=0.05cm] (ns4) at (8.5,2.2){};
        \node[shape=circle,draw=black, anchor=center,minimum size=0.05cm] (ns5) at (9,2.2){};
  \node[shape=circle,draw=black, anchor=center, minimum size=0.05cm] (nl1) at (7.5,3.2){} ;
   \node[shape=circle,draw=black, anchor=center, minimum size=0.05cm] (nl2) at (8,3.2){} ;
 \node[shape=circle,draw=black, anchor=center, minimum size=0.05cm] (nl3) at (8.5,3.2){} ;

\draw[->] (i_1) -- (nf1);
\draw[->] (i_2) -- (nf2);
\draw[->] (i_3) -- (nf3);

\draw[-] (nf1)--(ns1);
\draw[-] (nf1)--(ns2);
\draw[-] (nf1)--(ns3);
\draw[-] (nf1)--(ns4);
\draw[-] (nf1)--(ns5);

\draw[-] (nf2)--(ns1);
\draw[-] (nf2)--(ns2);
\draw[-] (nf2)--(ns3);
\draw[-] (nf2)--(ns4);
\draw[-] (nf2)--(ns5);

\draw[-] (nf3)--(ns1);
\draw[-] (nf3)--(ns2);
\draw[-] (nf3)--(ns3);
\draw[-] (nf3)--(ns4);
\draw[-] (nf3)--(ns5);

\draw[-] (ns1)--(nl1);
\draw[-] (ns1)--(nl2);
\draw[-] (ns1)--(nl3);

\draw[-] (ns2)--(nl1);
\draw[-] (ns2)--(nl2);
\draw[-] (ns2)--(nl3);

\draw[-] (ns3)--(nl1);
\draw[-] (ns3)--(nl2);
\draw[-] (ns3)--(nl3);

\draw[-] (ns4)--(nl1);
\draw[-] (ns4)--(nl2);
\draw[-] (ns4)--(nl3);

\draw[-] (ns5)--(nl1);
\draw[-] (ns5)--(nl2);
\draw[-] (ns5)--(nl3);

\draw[->] (nl1)--(q5);

\coordinate (t) at (8,4.4);

\draw[->] (nl2)--(t);

\draw[->] (nl3)--(q4);

\coordinate (t1) at (6,0.5);
\coordinate (t2) at (10,0.5);

\draw[-] (t1)--(i_1);
\draw[-] (t2)--(i_3);

\end{tikzpicture}
\caption{Illustration of the general structure of a soft first-order encoding for the sentence $\forall x:(R_1(x,y) \wedge \neg R_2(y)) \rightarrow R_2(x)$. The neural network maps assignments to variables $x$ and $y$ to the predicates $R_1$ and $R_2$, grounded by those variables. From the truth values of those predicates, the truth value of more complicated logical expressions can be computed iteratively using fuzzy or probabilistic interpretations of the operators. The result is used as a loss function to update the parameters of the neural network.}
\label{IDAT_ex}
\end{figure}
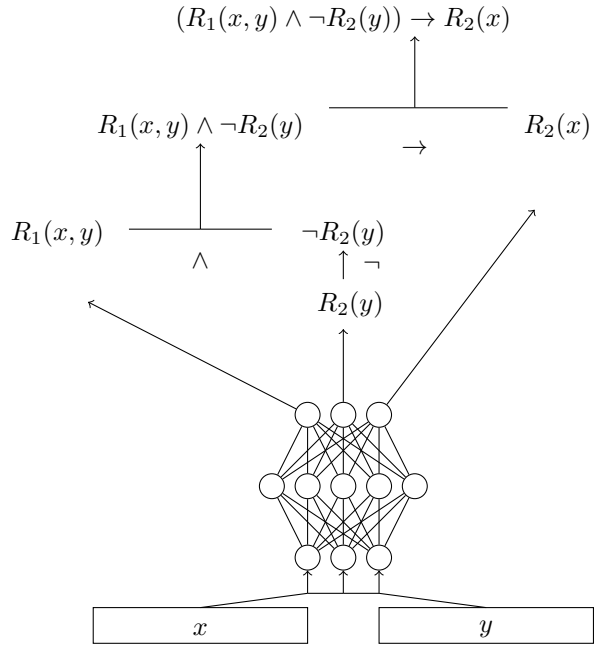

The neurons-as-atoms approach described in the previous section provides an intuitive way of encoding propositional logic into neural networks. When it comes to first-order logic, however, it is limited by the fact that the number of neurons required to model the logic may be infinite. In first-order logic, atoms are implicitly defined by predicates and variables. Each predicate represents a set of ground atoms defined by each possible variable assignment for the predicate. In full first-order logic, the domain of variables can have any cardinality, meaning that the set of atoms is infinite and thus cannot be represented by a neural network using the neurons-as-atoms paradigm. Even in cases where the domain of variables is restricted, the number of possible ground atoms can be prohibitively large. When translating datasets into knowledge bases, the number of  assignments for the variables is equal to the number of possible input values which rules out a neurons-as-atoms approach. To address this problem, most modern techniques use a distributed representation for atoms. That is, each atom is encoded by a specific activation pattern (embedding) in the neural network. This allows for a potentially infinite number of atoms to be represented and more closely aligns the logic being represented with the structure of the dataset. Before we describe the distributed atoms approach, we review the semantics of first-order logic.\\\\
The language of first-order logic is constructed from a set of predicates $\{P_1,P_2,...\}$, functions $\{f_1,f_2,...\}$, variables $\{x_1,x_2,..\}$, constants $\{c_1,c_2,...\}$, the logical symbols of propositional logic and the quantifiers $\forall$ and $\exists$. Each predicate and function has an associated natural number known as its arity representing the number of arguments it takes. A \textit{term} is a variable, a constant or a function of variables and constants. Formulas are constructed recursively in the following way:
\begin{itemize}
    \item If $P$ is a predicate of arity $n$ and $t_1,t_2,...,t_n$ are terms then $P(t_1,t_2,...,t_n)$ is a formula, in particular this is an atomic formula;
    \item If $\phi$ and $\psi$ are formulas then all combinations of $\phi$ and $\psi$ using the logical symbols of propositional logic are formulas;
    \item If $\phi$ is a formula containing a variable $x$ then $\forall x:\phi$ and $\exists x:\phi$ are formulas. In these resulting formulas the variable $x$ is said to be \textit{bound}.
\end{itemize}
A variable, $x$, in a formula, $\phi$ is considered to be \textit{free} if it is not bound. For simplicity we assume that there are an infinite number of variables which do not get reused for different quantifiers (i.e. if $x$ is a variable then it appears in the scope of at most one quantifier) but more sophisticated formulations can reuse the same variable for multiple quantifiers in a formula and make use of variable substitution for the purpose of reasoning. A sentence in first-order logic is one in which there are no free variables. The interpretations of first-order logic are known as first-order structures. These consist of the following:
\begin{itemize}
    \item A set, $D$, known as the domain;
    \item For each constant symbol in the language, $c_i$, an element $d_{c_i}\in D$;
    \item For each function symbol, $f_i$, a function $f_{M,i}:D^k\rightarrow D$ where $k$ is the arity of $f_i$;
    \item For each predicate in the language, $P_i$, a relation $P_{M,i}:D^K\rightarrow \{0,1\}$.
\end{itemize}
Given a first-order structure, $M$, define a variable assignment, $\mu$, as a function that assigns each variable in the language to an element in the domain. A variable assignment extends to assign domain elements to all terms in the language by $\mu(f_i(t_1,t_2,...,t_n))=f_{M,i}(\mu(t_1),\mu(t_2),...,\mu(t_n))$ and truth values to atomic formulas by $\mu(P_i(t_1,t_2,...,t_n))=P_{M,i}(\mu(t_1),\mu(t_2),...,\mu(t_n))$. Variable assignments can be used to define the truth value of general formulas in the following way: 
\begin{itemize}
    \item If $\phi$ and $\psi$ are sentences then the truth values of $\phi\wedge\psi$, $\phi\vee \psi$, $\neg\phi$, $\phi\rightarrow \psi$ and $\phi\leftrightarrow \psi$ are determined as done for propositional logic;
    \item If $\phi$ is a sentence with free variable $x$ then $\forall x:\phi(x)$ is true if $\phi(x)$ is true under all variable assignments which possibly differ from $\mu$ only in their choice of value for $x$ and $\exists x:\phi(x)$ is true if $\phi(x)$ is true under some variable assignment which differs from $\mu$ at most in its choice of $x$. 
\end{itemize}
Given a first-order sentence, $\phi$, and an interpretation, $M$, all variable assignments will give the same truth value for $\phi$ meaning that interpretations directly define the truth value of sentences. Given a first-order knowledge base $L$ (i.e. a set of first-order sentences), an interpretation, $M$, is a model of $L$ if it evaluates each sentence in $L$ to $True$.\\\\
We must also look briefly at fuzzy and probabilistic logics. Fuzzy first-order or propositional logic is similar to its non-fuzzy counterpart with the exception that the truth values are no longer restricted to $\{0,1\}$ but can take any value in $[0,1]$. The idea is that certain concepts have degrees of truth. Saying ``it's hot" is not necessarily true or false because everyone has a different threshold for what hot is. However, what can be said is that it becomes less true as the temperature goes down and more true as it goes up. The semantics of fuzzy logic are conceived in a similar way to propositional and first-order logic, except that logical symbols and quantifiers now define functions in $[0,1]$ instead of $\{0,1\}$. There are several ways of doing this resulting in various families of fuzzy logic. In particular, conjunction, $\wedge$, is most often replaced by a \textit{t-norm}, a function $\wedge:[0,1]\times[0,1]\rightarrow [0,1]$ satisfying the following:
\begin{itemize}
    \item Commutativity: $A\wedge B=B\wedge A$
    \item Monotonicity: $A\wedge B \leq C\wedge D$ if $A\leq C$ and $B\leq D$
    \item Associativity: $A\wedge (B\wedge C)=(A\wedge B )\wedge C$
    \item Identity $A\wedge 1=A$
\end{itemize}
Examples of \textit{t-norms} include $\min(A,B),A\cdot B,\max(0,A+B-1)$. Pairing a \textit{t-norm} with negation, defined by $\neg A=1-A$ implicitly defines the truth values of the other logic operators through the identities $A\vee B=\neg(\neg A\wedge \neg B)$ and $A\rightarrow B=\neg A \vee B$. Moving to first-order fuzzy logic, the first-order structures can be \textit{fuzzified} simply by interpreting the predicates of a language as functions with co-domain $[0,1]$ instead of $\{0,1\}$, i.e. $P_{M,i}:D^K\rightarrow [0,1]$. The semantic structure is otherwise unchanged. The universal and existential quantifiers are interpreted as the \textit{infimum} (greatest lower bound) and \textit{supremum} (least upper bound), respectively. This is intuitive as the degree to which `for all $x$' holds true is determined by the case where it least holds true, while the degree to which `there exists $x$' holds true is determined by the case where it most holds true.
\\\\
Similarly to fuzzy logic, probabilistic logic replaces binary truth values with values in the range $[0,1]$. The difference is that this value now represents the probability of a ground atom being true rather than a fuzzy truth value. In fuzzy logic, the world is assumed to be vague and it is modeled as such (with a range of values of how true \textit{hot} is). In probabilistic logic, the world is deemed to be crisp (it is either \textit{hot} or \textit{not hot}) but with probability $p$ assigned to the possible world in which it is hot and probability $1-p$ assigned to the remaining collection of possible worlds. In practice, probabilistic operators are defined to replace the standard propositional operators and first-order quantifiers \cite{probabilisticlogic}, as done for fuzzy logic using \textit{t-norms}. Differently from fuzzy logic's various \textit{t-norms}, probability theory offers a canonical formalization that is not a matter of choice, although probabilistic logic can also be viewed as defining a distribution over the interpretations of regular propositional logic (with many options for the choice of distribution). Computationally, fuzzy logic is more efficient because the probabilistic approach has to deal with a combinatorial explosion as the number of possible worlds increases. \\\\
\noindent In order to encode first-order fuzzy or probabilistic logic into neural networks, most modern neurosymbolic techniques use a distributed-atoms approach in which some neurons are used to represent atomic formulas and other neurons the arguments of predicates. Take a first-order predicate, $P(f_1(x),y)$. In the distributed-atoms approach, variables $x$ and $y$ are each assigned a fixed number of input neurons. Then additional neurons for $f_1(x)$ are added with connections from the neurons of $x$ to the neurons of $f_1(x)$. The neurons of $f_1(x)$ and $y$ are connected to the \textit{single} neuron representing $P(f_1(x),y)$. We may want to include hidden neurons as additional layers connecting the input variables/neurons to the output predicate/neuron. Under this formulation, the truth value of $P(f_1(x),y)$ is determined by the value of the corresponding neuron. This way an infinite number of ground atoms can be represented in a network by its input values. This also connects the dataset to the semantics of the logic directly by defining the domain of the first-order structures to be exactly the possible input values of the network. Refer to Figure \ref{IDAT_ex} for a simple visual example. Suppose, for example, that we are training a network on pairs of MNIST digits to output \textit{true} if the first digit is smaller than the second. We can translate this to first-order logic using the predicate $<$ and the first-order sentence $<(x,y)$, where the variables $x$ and $y$ range over grey-scale images of size $28\times28$. Given input images to the network, the final state of the network represents a prediction as to whether or not the first image depicts a smaller number than the second, or, in our first-order language, the network's output defines the truth value of $<(\mu(x),\mu(y))$, where $\mu$ is the variable assignment that assigns variables $x$ and $y$ to the pixel values of the first and second images, respectively. \\\\In the previous section we used the notion of an encoding function to define how the state of a neural network corresponds to interpretations of propositional logic. Now, the state gives us information about what truth-value a predicate (e.g. $<$) should take given the values of its arguments (e.g. $\mu(x)$ and $\mu(y)$). It does not give us any information about the truth value of other possible groundings that may not be in the data. Therefore, we say that the encoding mapping, $i$, maps to the set of interpretations in which the truth value of $<_M(\mu(x),\mu(y))$ is equal to the value of the corresponding neuron in the neural network. If we look at the set of \textit{all} input-output pairs for the neural network, it will give us the truth values of $<(\mu(x),\mu(y))$, for every possible variable assignment $\mu$. If we use the encoding mapping for each assignment, it gives us the collection of sets in which $<(\mu(x),\mu(y))$ is fixed for a particular value of $\mu(x)$ and $\mu(y)$ but arbitrary for every other possible value. If we take the intersection of all of these sets, it will consist of a single interpretation in which, for all possible values of $\mu(x)$ and $\mu(y)$, the truth value of $<(\mu(x),\mu(y))$ is equal to the output of the network given inputs $\mu(x)$ and $\mu(y)$. This is the interpretation that the neural network encodes. Unlike in the neurons-as-atoms case, in which each fixed state of the network gave us interpretations believed in by the network, now each state only gives us information about a given number of atom groundings. The beliefs of the network are represented by combining the information in \textit{all} of those states. Variants of this approach using fuzzy, probabilistic and fuzzy-probabilistic approaches have been developed in \cite{LTNfinal, diff_fuzz, deepproblog, deeplogicmodels, NeSYStar}. A generalization for many of the variants of this encoding technique has been put forward in \cite{deeplog}, in which the various logical operators can be interpreted in different ways. Combinations of fuzzy and probabilistic logic circuits then provide an error signal to the network to encourage it to learn outputs that conform to the desired logical circuit. This gives a template for the soft encoding of first-order logic: a first-order sentence is chosen to regularize the network, it is decomposed into its constituent operations, those operations are given fuzzy or probabilistic interpretations, a circuit is built or a loss function is defined (the circuit produces a loss signal to the neural network) to compute the truth value of the sentence having ground predicates as input, and finally a neural network is used to predict those predicates.\footnote{In \cite{BeneGarcez}, using the approach in \cite{LTNfinal} with concept activation values probing, a further step of symbolic knowledge extraction is added to this list. It is an example of the application of the neurosymbolic cycle. Interestingly, descriptions in first-order logic are extracted from the trained network, even though the network can only be trained with a finite number of ground atoms. The first-order descriptions, therefore, offer an extrapolation to infinite domains, which illustrates the potential value of the changing of representation, from neural to symbolic and vice-versa, in the neurosymbolic approach. Later, we will discuss the importance of measuring how close an extracted symbolic description may be to the trained network.} Figure \ref{deeplog} illustrates this. Soft encoding of this nature can also be used for logical inference. The trained network represents a set of interpretations which, along with a distance function on the set of interpretations (see Section 3.3), defines a belief function measuring the degree to which the neural network believes that a particular interpretation is true. Each interpretation satisfies a sentence to a certain degree ($0$ or $1$ for standard first-order logic or some value in $[0,1]$ for probabilistic and fuzzy logic). By aggregating the values over all interpretations (e.g. multiplying the values) we arrive at a truth value for the sentence in question. This is a particular case of the more general neurosymbolic inference procedure defined in \cite{nesyinference}. \\\\
  \begin{figure}
\begin{tikzpicture}[scale=0.8]
\centering
\useasboundingbox (0,0) rectangle (4,4);

 \node[shape=circle,draw=black, anchor=center,minimum size=0.8cm] (a) at (2.5,0.5){$x_{y_1,1}$};

  \node[shape=circle,draw=black, anchor=center,minimum size=0.8cm] (b) at (4,0.5){$x_{y_1,2}$} ;

   \node[shape=circle,draw=black, anchor=center,minimum size=0.8cm] (c) at (7.5,0.5){$x_{y_2,1}$};

  \node[shape=circle,draw=black, anchor=center,minimum size=0.8cm] (d) at (9,0.5){$x_{y_2,2}$} ;

  \node[shape=circle,draw=black, anchor=center,minimum size=0.8cm] (e) at (5.5,3.5){$x_{Ry_1y_2}$} ;

    \node[anchor=center,minimum size=0.8cm,label={[align=left]$\{M|R((x_{y_1,1},x_{y_1,2}),(x_{y_2,1},x_{y_2,2}))$\\$=x_{Ry_1y_2}\}$}] (f) at (13,2){} ;
%$\{M|R((x_{y_1,1},x_{y_1,2}),(x_{y_2,1},x_{y_2,2}))$\\$=x_{Ry_1y_2}\}$

    %$    \node[anchor=center,minimum size=0.8cm] (g) at (7,1){$(1,2,2,2)$} ;

  \coordinate (b_1) at (2.5,3);
  \coordinate (b_2) at (8,3.5);
  \coordinate (b_3) at (8.5,3.5);
  %\coordinate (b_3) at (1,5);
  %\coordinate (b_4) at (4,1);

%\draw[->] (b) -- (h2) node[near center, below, xshift=0.2cm,yshift=-0.9cm]{$1$};

  %\draw[->] (a) .. controls (2,1.75) .. (a)  node[near end, below,yshift = -0.1cm,xshift=-0.8cm]{$1$};

 \draw[->] (a) -- (e) ; 
 \draw[->] (b) --  (e) ;
  \draw[->] (c) --  (e) ;
    \draw[->] (d) --  (e) ;
    \draw[->] (b_2) --(b_3);

  %\draw[->] (f) -- (g) node[near start, below,xshift=0.3cm]{$h$};
  %\draw[->] (e) -- (f) node[near end, above,xshift=1.5cm]{};

  %\draw[->] (b_4) -- (b) node[near start, below,xshift=-0.0cm]{$-1$};

\end{tikzpicture}
\caption{Illustration of how an atom is represented in a neural network under the distributed-atom encoding. If $R(y_1,y_2)$ is a predicate in a first-order language whose interpretations are in the domain $\mathbb{R}^2$ then the variables $y_1$ and $y_2$ are represented by two input neurons each, while $R(y_1,y_2)$ is represented by an output neuron with activation value in $[0,1]$. We use $R$ because in essence a logical predicate implements a relation. The states of the network map to the set of interpretations in which the atom formed from grounding $R(y_1,y_2)$ with the values of the input neurons is given truth value $x_{Ry_1y_2}$. Taking the intersection over all input values leaves the network representing interpretations in which all groundings of $R$ are assigned a specific truth value determined by the network.}
\label{deeplog}
\end{figure}
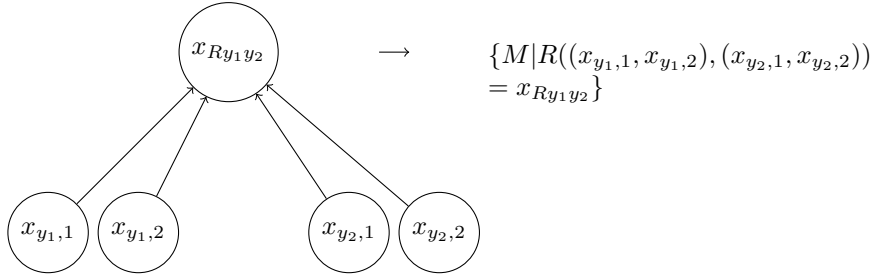

%\forall x,y: R_1(x,y)\wedge \neg R_2(y)\rightarrow R_
\subsection{Other Logics and Encodings}
While first-order and propositional logic have captured most of the attention for neurosymbolic encoding, there are other logics with rich theory and important applications that may be valuable to encode in a neural network. In particular, modal logic is a large area of study in computer science with applications in robotics, classical AI and systems verification, where temporal logic is a dominant formalism. Modal logic extends propositional and first-order logic with the operators $\square$ and $\diamondsuit$ representing necessity and possibility respectively. The semantics of Modal logic involve a set of possible worlds which are related via an accessibility relation. Given a possible world, $w$, the statement $\diamondsuit P$ is true if $P$ is true in some possible world, $u$, that is accessible from $w$; $\square P$ is true in $w$ if $P$ is true in all worlds that are accessible from $w$. Many variants of Modal logic, such as Temporal, Epistemic, and Deontic logic exist, each with a different focus and target applications. The encoding of Modal logic programs in neural networks was first studied in \cite{Nesybook1}, where the accessibility relation is pre-defined and used to create multiple networks (agents) that communicate according to the definitions of $\square$ and $\diamondsuit$, each network encoding a logic program using CILP. However, while there is now a well-defined approach to encoding first-order and propositional logic, how best to define an encoding function for the many flavors and applications of Modal logic is a more open question, especially in the case where the accessibility relation may be learnable. This area of research requires more attention before a consensus can be arrived at in terms of how the encoding function should be defined.\\\\
While there were effort towards encoding Modal logic in neural networks, we are not aware of attempts to encode \textit{higher-order} logic formally in neural networks. Higher-order logic is a generalization of first-order logic in which quantification can range not only over the objects of the domain but over sets of objects, and sets of sets of objects, etc. Second-order logic, for example, allows quantification over predicates themselves. The important point that we would like to make here is that, despite the existence of useful encoding techniques for propositional and first-order logic, the problem of semantic encodings is far from solved. For this reason, in order to study semantic encodings in general, we introduced a generic framework which describes the overall process of a semantic encoding given an encoding function. Having this is expected to help organize the study of semantic encodings by making clear the structure of the encodings as they have been conceived in the literature without having to go into the details about specific encodings. In the case of higher-order logic, for example, semantic encodings should be studied within our framework in connection with the recent efforts in neurosymbolic program synthesis and the integration of functional programming and neural networks.

\section{Semantics of Neural Encodings}
%\subsection{Definition of a Neural Encoding}
We have seen how many neural encodings use the semantics of a logic to encode knowledge into a neural network. We have also seen that the way in which this is done varies depending on the type of logic and neural network. If we want to analyze the overall utility of neural encodings, we need to be able to give a general definition for how to encode semantics within a neural network. Without a definition, one can point for example to research on reasoning shortcuts as a problem of neurosymbolic AI \cite{marconato2023neurosymbolic}, but the analysis will be limited to the results obtained by the specific encoding techniques used in experimental evaluations. With a definition, it should be possible to generalize results like reasoning shortcuts results based on the analysis of encodings from first-principles.\\\\
Luckily, despite the differences displayed by the various encoding techniques seen in the previous section (and the countless other techniques in the literature), they all share some common elements, allowing us to give a general definition. All semantic encodings are defined by three elements, as follows.\\\\
\textbf{The Encoding Map ($I$):} The most crucial component that all encodings share is an encoding map, $i$, which maps states of the neural network to the interpretations of a logical system. How they do this depends on the technique, but as we saw there are generally two approaches: for propositional logic, each neuron is assigned a propositional atom and the value of that neuron determines the truth value of the atom for the corresponding interpretation. We use $I_{NAT}$ (neurons-as-atoms) to denote an encoding of this form. For first-order logic, some neurons represent predicates whereas others represent the variables used to ground those predicates. The values of the neurons representing variables determine which set of ground atoms the state represents. The values of the neurons representing predicates determine the truth-values of those ground predicates. We use $I_{DAT}$ (distributed-atoms) to denote an encoding of this form. While these are the most common forms of encoding, there are other ways of representing interpretations in a neural network, some of which we saw and some that no doubt are yet to be developed. For this reason, we do not want to restrict our encodings entirely to $I_{NAT}$ and $I_{DAT}$. However, as we will discuss later, allowing for a completely arbitrary encoding function will allow any neural network to encode any knowledge base, something that we will see make encoding meaningless. For now, we will assume that there is some set of encoding maps, $I$, that can meaningfully represent the interpretations of a logic in a neural network. It is also important to note that in practice neural networks have hidden units which are added to increase computational power but are not intended to have any semantic interpretation (although it may be possible to derive an interpretation from neighboring neurons). Thus, $I$ will be a map from \textit{visible} neurons to interpretations where the values of the hidden neurons have no effect on the interpretation mapped to by $I$.\\\\
\textbf{The Stable States ($X_{inf}$)}: The next question we have to answer is \textit{which states of the neural network represent important information}? An encoding map associates \textit{every} state of the neural network to a set of interpretations, but in the previous section we saw that it is only specific states of the neural network that are of interest. For feed-forward networks, these states are the input/output pairs; for recurrent networks they are the stable states. We therefore define a set $X_{inf}$ as the set of states of interest. Formally, we define this set to be the set of all states $x\in X$ such that there exists a sequence of positive integers $\{t_i\}_{i=0}^{\infty}$ such that for $\{N^{t_i}(x_0)\}_{i=1}^{\infty}$, $t_i<t_{i+1}$, we have that $\lim\limits_{i\rightarrow \infty} N^{t_i}(x_0)=x$, remembering that $N$ represents the function updating the state of the neural network. This set represents the set of states that the network converges to over time. It is technically more general than the cases we described as it allows for a neural network to converge to an infinite cycle instead of a single point as was the case for the network shown in Figure \ref{fig:recurrent_net}. This captures all our desired cases while leaving space for future encoding techniques to make use of networks with more complicated dynamics. In any case, we require our networks to be stable, that is, for all initial states $x\in X$, $\lim\limits_{t\rightarrow \infty} N^t(x)\in X_{inf}$. We assume that, given enough time, the network will converge to a state in $X_{inf}$. We do not consider encodings that involve networks which diverge or exhibit chaotic behavior.\\\\

\noindent \textbf{The Aggregation Function ($Agg$):} Finally, given the states of interest, $X_{inf}$ and a map from states to sets of interpretations, we can form a set $i(X_{inf})$ consisting of all sets of interpretations associated with the stable states. The beliefs of our network should be represented in the form of a set of interpretations, not a set of a set of interpretations, so we need a way of aggregating the beliefs represented by each state in $X_{inf}$ to produce a final set of beliefs. In our propositional examples, each state in $X_{inf}$ represented an assignment of truth values to all relevant atoms. Therefore, each state represented models of the knowledge base. To collect all of the models together we simply took the union over these sets. In our examples with first-order logic, on the other hand, each state represented the desired truth assignments to only a subset of the ground atoms that we were interested in. Because each state in $X_{inf}$ mapped to the set of models in which a particular set of ground atoms were given truth values, in order to retrieve the set of interpretations that assigns all ground atoms to their truth values, we must take the intersection over these sets. Again, to keep the definition general, we allow for an arbitrary aggregation function. We simply specify that there is some function $Agg:2^{2^\mathcal{M}}\rightarrow 2^{\mathcal{M}}$ that combines all the beliefs represented by stable states into a final set of beliefs for the network. All together, this gives us the following definition.

\begin{definition}
\label{semanticneuralmodel}
\textit{Suppose we have a neural network, $N$, with state space $X$ and a logical system with language $\mathcal{L}$ and interpretations $\mathcal{M}$. If $L$ is a knowledge base in the language then, given an encoding function $i\in I$, $i:X\rightarrow 2^{\mathcal{M}}$, and an aggregation function, $Agg:2^{2^{\mathcal{M}}}\rightarrow 2^{\mathcal{M}}$, we define:
\begin{itemize}
    \item Let $\mathcal{M}_N=Agg(\{i(x) | x\in X_{\textit{inf}}\})$ and let $\mathcal{M}_L$ be the set of models of $L$. $N$ is called a \emph{neural model} of $L$ under $I$ and $Agg$ if $\emptyset \subset \mathcal{M}_N \subseteq \mathcal{M}_L$.
    \item $N$ is called a \emph{semantic encoding} of $L$ under $I$ and $Agg$ if it is a neural model of $L$ under $I$ and $Agg$ and $L\vDash L'$ if and only if $\mathcal{M}_N\subseteq \mathcal{M}_{L'}$.
    \item A set of neural networks, $\mathcal{N}$, and the logical system, $\mathcal{S}$, are \emph{semantically equivalent} under $I$ and $Agg$ if every knowledge-base of $\mathcal{S}$ has a semantic encoding under $I$ and $Agg$ into a neural network in $\mathcal{N}$ and all neural networks in $\mathcal{N}$ are semantic encodings under $I$ and $Agg$ of some knowledge-base $L$ in $\mathcal{S}$.
\end{itemize}
}
\end{definition}

\begin{figure}
    \centering
    \includegraphics[width=0.7\linewidth]{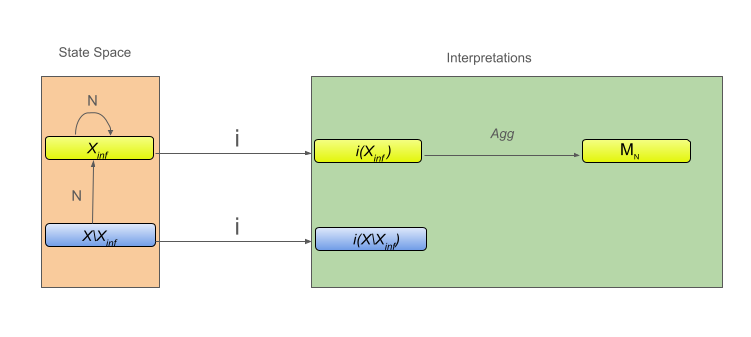}
    \caption{A visual representation of a semantic encoding. $N$ represents a (possibly multi-step) update of the state of the neural network. By repeatedly updating, the state will converge eventually to a state in $X_{inf}$. Each of those states is mapped to a set of interpretations by $i$ and the collection of these sets is aggregated using $Agg$ into a final set of interpretations representing the beliefs of the neural network, $\mathcal{M}_N$.} 
    \label{fig:placeholder}
\end{figure}
\noindent Notice that $\vDash$ represents logical entailment: we say that $L\vDash L'$ if, for all $M\in \mathcal{M}$, if $M$ is a model of $L$ then $M$ is a model of $L'$. In practice, we usually replace this condition with the simpler condition $\mathcal{M}_N=\mathcal{M}_L$. Figure \ref{fig:placeholder} shows the general structure of a semantic encoding. Intuitively, each state of the network represents a set of beliefs. We can think of the set of interpretations mapped to by $i$ as the possible worlds in which these beliefs are satisfied. As the network updates itself, it revises a set of beliefs. The network reaching a stable state represents arriving at a set of beliefs that it considers to be true. The total set of beliefs of the network is simply the aggregation of all of these belief sets. Consider again the example shown in Figure \ref{fig:recurrent_net} and Figure \ref{fig:state_transition}. The stable states of this network are $(0,0)$ and $(1,1)$ representing the sets of interpretations in which $A=B=False$ and $A=B=True$ respectively. With $\cup$ as an aggregation function, we get that $\mathcal{M}_N$ represents the set of interpretations in which either $A=B=False$ or $A=B=True$. These are exactly the models of $(A\wedge B) \vee (\neg A\wedge \neg B)$ meaning that $N$ encodes the knowledge base consisting of this sentence. Notice that this definition unites both the neurons-as-atoms approach and the distributed-atoms approach. In both cases, an encoding function is used to determine interpretations from network states and the encodings of the stable states are aggregated to produce the final set of interpretations representing the network's beliefs. Having this general framework lets us reflect on the role that various components may play in encoding semantic information into neural networks. For example, adopting $\cap$ as aggregation in the above case would produce an inconsistency. The problem of determining what makes an encoding map valuable or appropriate is reflected upon in the Philosophy of Mind problem of individuating computational systems being implemented by physical ones as we will discuss now.

\subsection{Encoding Maps and the Wordstar Problem}
The central component of a semantic encoding is the encoding map. This determines how interpretations are represented in the neural network. We gave two examples representing the standard way of encoding propositional and first-order logics, $I_{NAT}$ and $I_{DAT}$, respectively, but in principle the encoding function could be anything. This presents a problem because an arbitrarily defined encoding map could encode virtually any logic into any neural network. For example, suppose we have some knowledge base, $L$ in some logic with models $M_L$, and a neural network with a single stable state, $x_{inf}$. We can define an encoding map $i_L$ with $i_L(x_{inf})=M_L$ where $i_L(x)$ is defined arbitrarily for $x\neq x_{inf}$. Allowing for arbitrary encoding maps clearly makes the notion of a semantic encoding meaningless as the same network can be thought of as encoding every knowledge-base of every possible logic depending on how it is encoded. This is why putting restrictions on the encoding map is essential. This same problem appears in the philosophy of mind as a challenge to the computational theory of mind. \\\\ The basic idea of the computational theory of mind is that cognition is at least in part a computational process and the human mind can be thought of as an information processing machine implementing specific computations \cite{computationaltheory}. The challenge to this comes from Searle's Wordstar metaphor \cite{wordstar}. For the same reasons that apply to semantic encodings, every physical system can be seen as implementing any computation if you are allowed to arbitrarily define how states of the physical system correspond to components of a computational system. You can see a direct example of this in \cite{indeterminacy} where a simple network is proposed that implements both the \textbf{OR} function and the \textbf{AND} function depending on how you group the states together. Searle made this point by stating that the wall behind him implements the Wordstar program and, indeed, if it is large enough, any program. Thus, the fact that the human brain can be seen as implementing some computational system is not sufficient to differentiate it from any other physical object and is therefore not a suitable explanation of cognition.\\\\
Rebuttals to this take the obvious approach of clarifying that it is not that the brain implements a computational system, but a particular computational system that relates to the physical world in a particular way. In these rebuttals, arbitrary associations of physical states are considered invalid. For instance, in the semantic view, the important physical states carry semantic information about the outside world \cite{semanticview}, in the mechanistic view, a more granular look at the physical system is taken where components of the physical system have to relate to each other in a way that somehow preserves the relationships between components of the computational system \cite{mechanisticview}. An argument of particular interest is that the encoded computational system has to have some sort of benefit to the organism \cite{teleoview}. After all, the encoding of Wordstar in an organism's brain bears no relation to its interaction with the environment. Instead, a mixture of evolutionary and social factors ground perceptual input in a larger system that produces rewards for that organism. Computational structures that process information in a way that helps an organism achieve some external goal provide more compelling descriptions of the brain's function than arbitrary ones. Similarly, semantic encodings that have a bearing on the ability of a network to learn a dataset are more interesting to us than arbitrary semantic encodings. The fact that we have a specific practical application for semantic encodings gives us clear conditions under which a semantic encoding is meaningful or relevant. The encodings we have shown so far get around this problem by ensuring that the structure of the network is closely aligned with the structure of the logic. In the case of $I_{DAT}$, the logic is specifically designed to resemble the dataset to the point that, as we will see, the dataset itself can be considered a knowledge base of the logic. For this reason, most modern encodings avoid this approach entirely, however, a general set of conditions under which an encoding function is relevant to the dataset has yet to be produced. It would be beneficial to find explicit criteria that an encoding function has to satisfy given a dataset and learning algorithm in order for that encoding function to encode information about the dataset that is relevant to the network under the learning algorithm. Having such conditions could serve as a guide to the development of future encoding techniques, providing a path to the analysis of the impact of the encoding of specific knowledge bases in an abstract setting. For the moment, we will specifically examine how $I_{DAT}$ encodes typical neural network datasets and how this encoding can potentially be used to model the impact that the encoding of various knowledge bases may have on the learning process. Before that, however, we briefly discuss what it means to approximately encode a knowledge base.
\subsection{Approximate Encodings}
While hard encodings ensure that the neural network encodes a knowledge base exactly, in practice we are often presented with data that contradicts our background knowledge. This is especially true in the case of fuzzy or probabilistic logic. Soft encodings will weigh the knowledge base against the data to produce a neural network that does not necessarily fully align with either. However, we still expect that neural network to come closer to satisfying the knowledge base than a network trained solely on the data. For this purpose we need a way of measuring how far a neural network with its encoding and aggregation functions may be from being a neural model of a knowledge base. As with most neurosymbolic encodings, this is generally done in an ad-hoc way, although the measures rely on fuzzy or probabilistic notions of distance. Instead, we will use the term \textit{fidelity}, taken from the literature on rule extraction \cite{DAVILAGARCEZ2001155}, to denote a function that measures how far a neural network is from modeling a knowledge base. 
\begin{definition}
\textit{Let $\mathcal{N}$ be a set of neural networks and suppose that $\mathcal{S}$ is a logical system with language $\mathcal{L}$ and interpretations $\mathcal{M}$. A function, $Fid$, whose input consists of triples $(N,i,Agg)$ and knowledge bases $L$ is a fidelity measure if $Fid((N,i,Agg),L)\in [0,1]$ and $Fid((N,i,Agg),L)=1$ if and only if $N$ is a neural model of $L$ under $i$ and $Agg$.}
\end{definition}
\noindent One way of defining a fidelity measure is to look at $\mathcal{M}_N$, the set of interpretations represented by $(N,i,Agg)$, and $\mathcal{M}_L$, the set of models of $L$. If we already have some distance function defined on $\mathcal{M}$ then we can use the Hausdorff distance $d(\mathcal{M}_N,\mathcal{M}_L)$ to define a fidelity measure. This is just the maximum of $\sup\limits_{M\in\mathcal{M}_N} (M,\mathcal{M}_L)$ and $\sup\limits_{M\in\mathcal{M}_L} (\mathcal{M}_N,M)$ or the longest possible distance for any interpretation in $\mathcal{M}_N$ and $\mathcal{M}_L$. Notice that we can always use the discrete distance measure on interpretations, $d(M_1,M_2)=0$ if $M_1= M_2$ and $1$ otherwise. With this distance function we can define a fidelity measure by $\frac{d_{max}-d(\mathcal{M}_N,\mathcal{M}_L)}{d_{max}}$, where $d_{max}$ is the maximum possible value of $d$ (if $d$ is unbounded, we can choose a specific $d_{max}$ as the maximum distance and set the numerator to $\max(0,d_{max}-d(\mathcal{M}_N,\mathcal{M}_L))$). Using the discrete distance function to create a fidelity measure will just result in the function $Fid((N,i,Agg),L)=0$ if $N$ is a neural model of $L$ under $i$ and $Agg$, and $1$ otherwise. Instead, most techniques use fuzzy or probabilistic methods of measuring distance, for instance:
\begin{definition}
    $Fid_{fuzzy}((N,i,Agg),L)=\inf\limits_{M\in \mathcal{M}_N} \SatAgg\limits_{[a,b]:\phi \in L} \{1-d(M(\phi),[a,b])\}$,
\end{definition}
\noindent which defines the distance to satisfying a fuzzy knowledge base consisting of sentences of the form $[a,b]:\phi$, where $0\leq a \leq b\leq 1$ are fuzzy values. A fuzzy interpretation is a model of this sentence if it assigns $\phi$ a fuzzy truth value in $[a,b]$ (remember that $\mathcal{M}_N$ is the set of interpretations represented by the network and $M(\phi)$ is the fuzzy truth value of $\phi$ given by interpretation $M$, and $d(M(\phi),[a,b])$ is the distance between the fuzzy value assigned by $M$ and the desired set of fuzzy values). SatAgg is some aggregation function, $\min$ works for most cases but there are other choices, such as multiplication. We can intuitively understand this fidelity as measuring the closest that the network interpretations will come to assigning fuzzy truth values in the desired range to each $\phi$.
\\\\
Another common measure applies specifically to networks which define probability distributions over a set of interpretations. When that is the case, we can measure how far the network is from defining an encoding of $L$ using: 
\begin{definition}
    $Fid_{prob}((N,i,Agg),L)=P(M\in\mathcal{M}_L)$, where $P$ is a probability distribution defined by $(N,i,Agg)$ over the interpretations of the language of $L$.
\end{definition}
\noindent For example, imagine we have a neural network with three output neurons representing propositional variables $Y_1,Y_2,Y_3$ under an encoding in $I_{NAT}$ and $\mathcal{M}_N$ consisting of a single probabilistic interpretation $Y_1=0.4,Y_2=0.6,Y_3=0.2$. If we have a propositional knowledge base $\{Y_1\wedge Y_2\wedge \neg Y_3\}$ then the probability that $N$ models $L$ is the probability that $Y_1=Y_2=1$ and $Y_3=0$ which is just $0.4\cdot 0.6\cdot 0.8=0.192$. In order to derive a probability distribution over interpretations from $(N,i,Agg)$, we simply need to define a distribution over $\mathcal{M}_N$, generally a uniform distribution will be sufficient but depending on which aggregation function we use the network itself might learn a distribution over this set. If interpretations $M_N \in \mathcal{M}_N$ define a probability distribution over some other set of interpretations (such as we just saw from the probabilistic interpretation of propositional logic to a standard binary interpretation) then $P(M)=P(M|M_N)P(M_N)$.\\\\
With this simple account of fidelity measures we conclude our description of semantic encodings. We can see that many of the problems facing neural encodings described in the opening sections can be better understood as properties of $i$, $Agg$, and $Fid$ and their relationship to a dataset and learning function. %While entirely general conditions relating these factors have not been described, we know that soft-encodings of first-order logic 
\section{Semantics of Deep Learning} 
\subsection{Learning as a Neurosymbolic Encoding Problem}
In machine learning, a classification task consists of finding a function that can accurately map input data, $x\in X$, to one or more labels $y\in Y$. It is assumed that each input data point has at least one corresponding label that accurately describes the data. For example, in image classification, input data consists of an image and the corresponding labels may be a subset of objects that appear in that image. A classification task, in other words, assumes the existence of a function $f:X\rightarrow 2^Y$ which maps each input data point to its set of labels. This defines a set, $X\times Y_X \subset X\times 2^Y$  by $X\times Y_X :=\{(x,f(x) | x\in X\}$.\\\\

\noindent The definition of a classification task is, thus, given a subset of $I_{train} \subset X\times Y_X$, approximate $f$. Using feed-forward neural-networks, a network $N$ is used to define a function $f_N:X\rightarrow 2^Y$ which can be compared to $f$. The output of $f_N$ on the input training data is compared to the actual labels given by training data. In other words, for $(x,Y_x)\in I_{train}$, $f_N(x)$ is computed (learned), and compared with $Y_x$. Algorithms such as gradient descent are used to update the parameters of $N$ in such a way as to bring $f_N(x)$ closer to $Y_x$. The performance of the resulting network is evaluated by measuring the average difference between $f_N(x)$ and $Y_x$ on a test set, $I_{test}\subset X\times Y_X , I_{test}\cap I_{train}=\emptyset$.\\\\ We can of course re-frame the learning task as a search for a first-order knowledge base. Define a first-order language with unary predicates, $\phi_y$ for each $y\in Y$ and no other predicates. Assume a constant $c_x=x$ exists for each $x\in X$. Define the interpretations of this language to be those first-order structures with domain $X$ in which each $c_x$ is assigned $x$. The training set can then be defined as a knowledge base, $L_{train}$ consisting of all those sentences of the form $\phi_y(c_x)$ where $(x,f(x))\in I_{train}$ and $y\in f(x)$. The goal is to infer, from the knowledge base $L_{train}$, the interpretation in which, for all $x\in X$, $\phi_y(x)$ is true iff $y\in f(x)$. From a purely logical point of view, this is impossible, as nothing in the training data will allow us to logically generalize to unseen examples. There is no reason to favor one model of $L_{train}$ over another. Neural networks and other learning methods exploit an assumption of continuity, that is, similar inputs will have similar labels. This notion of similarity is \textit{extra-logical} and relies on the statistical properties of the distribution of the input features and output labels. \\\\
What if, however, we had additional knowledge in our language? For example, what if we had a hierarchy of labels such that every time that a label is detected higher in the hierarchy, a label must also be detected lower in the hierarchy. This amounts to a partial order $<$ on $Y$ and can be encoded in the knowledge base, $L_{<}$, consisting of all sentences of the form $\phi_y(c_x)\rightarrow \phi_y'(x_c)$ with $y<y'$ (or, revisiting past examples of the softmax layer, labels could exclusionary give us sentences of the form $\phi_y(x)\rightarrow \neg \phi_{y'}(x)$). Other knowledge bases might make use of additional predicates or functions to describe the relationship between the input variables and the labels. Take for instance a predicate $rotate(x_1,x_2)$ which, for 2D images $x_1$ and $x_2$, is defined to be true if $x_1$ can be obtained by rotating $x_2$. Because rotation does not change which objects are found in an image, we might add to the knowledge base $L_{conv}:=\{\phi_y(x_{1_c}) \wedge rotate(x_1,x_2) \rightarrow \phi_y(x_{2_c}) | y\in Y x_1,x_1\in X\}$.\\\\
It is not difficult to see that when conceived in this way, a neural network learning a classification task is also learning a neural encoding of $L_{train}$ under $I_{DAT}$; the output neurons of the network represent the predicates $\phi_y$ and the input neurons represent their variables' arguments. The soft encoding techniques we described can thus be seen as generalizations of the regular learning problem. Beyond classification tasks, when learning regression tasks in which the output is in $[0,1]$, we can likewise interpret the problem as searching for a particular model of some kind of continuous logic, either fuzzy, probabilistic, or some combination thereof. If, for an input value, $x$, the output neurons give, for example, values $y_1=0.5,y_2=0.25$ then our knowledge base is $\{0.5: \phi_{y_1}(c_x),0.25: \phi_{y_2}(c_x)\}$. At the beginning of this section, we described two problems that needed to be addressed in order for neurosymbolic encodings to be useful, the first was representational, the way a neural network represented a logic must in some sense be consistent with how that logic describes the knowledge base. We can see in this example that encodings under $I_{DAT}$ satisfy this as the semantics of the logic are defined in terms of the dataset. The general question of representation, however, is not solved as there are other types of neural networks and logics for which a `natural' representation of semantics has yet to be developed, but for feed-forward neural networks representing first-order logic we can move on from the first problem facing encodings to the second, when is a knowledge base beneficial to learning? 
\subsection{When is Background Knowledge Beneficial?}
As previously outlined, a dataset used for a classification task can be expressed as a knowledge base in first-order logic and the process of learning that dataset can be reinterpreted as a search for models of the corresponding knowledge base.\\\\
The goal of semantic encoding is to find an additional knowledge base that will improve the chances that the model found by the network generalizes well to unseen data by satisfying the test set examples. When background knowledge is added, one wishes that the neural network will be more likely to generalize beyond the training examples. In this section, we show how the formalization of semantic encoding can be used to derive properties of the learning algorithm to explicitly measure the effect of background knowledge without reference to the particular encoding technique.\\\\

%Instead of deriving results for LTNs or DeepProbLog we give results that apply for all soft encodings using $I_{DAT}$. 
%The research in this direction is ongoing so many questions remain unanswered but it should serve to illustrate the importance of the underlying relationship between the semantics of a knowledge base and the network + algorithm attempting to learn that knowledge base.

First we assume that the set of network models is finite. Most classification problems are not formulated in this way since $X$ can be an arbitrary subset of $\mathbb{R}^n$. However, in practice, due to the finite memory capacity of computers, all neural models are technically finite as the input has a limited precision. We could expand our analysis to include the continuous case but for simplicity we will stick to the finite case. In the finite case, we have an input space $X\subset \mathbb{R}^n$ with $|X|<\infty$ and a set of labels $Y$, along with a set of training examples, $I_{train}$. We have seen how this defines a first-order language. The set of ground atoms in this language is $At=\{\phi_{y_j}(x_i)|y_j\in Y, x_i \in X\}$. An interpretation of this language is a map $M:At\rightarrow T$ where $T$ is a set of truth values; for now assume $T=\{0,1\}$. We can define a distance between interpretations $m_1$ and $m_2$ as:

\begin{equation}
d(M_1,M_2)=\frac{\sum\limits_{\phi_y(x)\in At} |M_1(\phi_y(x))-M_2(\phi_y(x))|}{|At|}, 
\end{equation}

\noindent which, with binary truth values, is just the percentage of atoms that are assigned the same truth values. This in turn lets us define a distance between an interpretation and a knowledge base as: $d(M,L)=\min\limits_{M_i:M_i\vDash L} d(M,M_i)$, where $M_i \vDash L$ indicates that $M_i$ is a model of $L$.\\\\
Assume we use a feed-forward neural network, $N$, to learn to label the set $X$ with labels in $Y$. The network encodes interpretations of the above logical system in the sense described in the previous section. In particular, a neural network assigns a truth value to every element in $At$ and thus we can measure the distance between the neural network and a knowledge base, $d(N,L)$, as the distance between the network's interpretation and the knowledge base. Assume that we have $I_{train}$,$I_{test}$, and $L_{train}$ as described before and a learning algorithm, $A$, that updates the parameters of the network $N$ using $L_{train}$. We will measure the effect of adding a knowledge base, $L$, as background knowledge, given some assumptions about the learning algorithm.\\\\
\textbf{Assumption 1}: Sufficient Capacity. Let $N_t$ be the network resulting from applying learning algorithm $A$ to network $N$ $t$ times with knowledge base $L$ (i.e. the network after $t$ epochs of training). $\lim\limits_{t\rightarrow \infty}d(N_t, L)=0$\\\\
This says that our learning algorithm is able to learn a knowledge base given enough time. When $L$ is derived from a training set this says that we are able to learn the training set with the given network. Usually this boils down to two properties: having enough neurons for the network to implement a function complex enough to recreate the input and output of the training set, and methods to prevent the learning algorithm from getting stuck in a local minima. In practice we do not expect the network to achieve $100\% $ accuracy on the training set, so we can relax the condition in our assumption so that the distance must be less than some small value $\epsilon$.\\\\
Given that we randomly initialize the network parameters along with the fact that most learning algorithms used today are stochastic in nature, the learning process under Assumption 1 creates a distribution over the models of $L$, $P(M)$, which is the probability that the learning algorithm converges to $M$ for a randomly initialized network after enough epochs.\footnote{If we do not expect the learning algorithm to land exactly on a model, this is just the probability that $M$ is the closest model to the network post-training. In the case of binary truth values this will be unique.}\\\\
Of these models, there is only one, $M_{true}$, that is the true model we are trying to find. We want the learning algorithm to result in $P(M_{true})$ being as large as possible. How does background knowledge contribute to this? Suppose we add knowledge base $L'$ and train our network on $L\cup L'$. By Assumption 1, this will induce a distribution on the models of $L\cup L'$, $P'(M)$. We look for conditions on $L'$ that result in $P'(M_{true})>P(M_{true})$. Intuitively, we should get an increase in probability when $L\cup L'$ has fewer models than $L$. This is because there are fewer possible points to converge to. When $\mathcal{M}_L\subseteq \mathcal{M}_{L'}$, the probability distribution over models should not change. This will be satisfied if our learning algorithm uses the gradient to minimize the distance between $N$ and the models of $L\cup\L'$; if the models of $L$ and $L \cup L'$ are identical then the distance between $N$ and the models of $L\cup L'$ is the same as the distance between $N$ and the models of $L$, meaning that the gradients will be identical. So as long as the error function reflects the distance between interpretations, adding $L'$ changes nothing about the learning process.\\\\
Assume instead that $|\mathcal{M}_{L'}\cap \mathcal{M}_{L}| < |\mathcal{M}_{L}|$. How much does this affect the probability of $M_{true}$? \\\\
\textbf{Property 1}: $P_{L\cup L'}(M)=P_{L}(M| M \vDash L')$.
\\\\
If property 1 is satisfied and $P(M\not\vDash L')\neq  0$, then it is easy to see that $P_{L\cup L'}(M')>P_{L}(M')$ for all $M'$ satisfying $L'$. By definition $P_{L\cup L'}(M)=\frac{P(M \wedge M\vDash L')}{P(M\vDash L')}$. Since we are choosing $L'$ to better describe the desired model, $M_{true}$, we know that $M_{true} \vDash L'$ meaning that $P_{L\cup L'}(M_{true})=\frac{P(M_{true})}{P(M\vDash L')}$. If $P(M\not\vDash L')\neq  0$ then $P(M\vDash L')=1-P(M\not\vDash L')<1$ so $P_{L\cup L'}(M_{true})>P_{L}(M_{true})$. Furthermore, the larger $P(M \not\vDash L')$ is, the larger $P_{L\cup L'}(M_{true})$ becomes. This makes intuitive sense as the more potential models we eliminate with $L'$ the fewer possible models there are for the network to converge to, and the likelihood that it converges to the correct one increases. This probability becomes $1$ if you choose $L'$ such that $M_{true}$ is the only model of $L\cup L'$.\\\\
If Property 1 is satisfied then all models of $L\cup L'$ increase their probability compared to $P_{L}$. So, the goal is also to choose $L'$ to be as small as possible, otherwise we could simply use $L'$ to specify $M_{true}$ exactly, in which case there would be no need to learn anything because we already know what the correct model is. %This means that adding a reasonable $L'$ to background knowledge could simply cause the network to converge to some other incorrect solution with high probability. 
Another, more fundamental issue with Property 1 is that the addition of supplementary background knowledge can change the shape of the loss function in unpredictable and complicated ways. It is very possible that the addition of $L'$ makes learning $M_{true}$ less likely in practice. The problem is that we do not know the prior likelihood of models from $P$ and if there are additional factors that make $P$ favor certain models over others. 
%then we could again end up in the same situation in which we essentially need to encode all of $M_{true}$ in the background knowledge before there is a decent chance that the network converges to it. 
Property 1 does, however, provide guideline for designing new neurosymbolic learning algorithms that would allow us to be confident that adding appropriate background knowledge in continual learning mode would increase the chances of learning $M_{true}$ and generalizing well. There is another property that can help with this. \\\\
\textbf{Property 2}: Low-Complexity bias.\\\\
%If neural networks learned solutions to the training set arbitrarily, they would have difficulty generalizing on any dataset as there would be no reason to favour one solution over another. 
It is thought that part of what makes neural networks effective learners is that they have an implicit bias towards less complicated solutions \cite{occamvnfl}. This is in fact an assumption of the whole of machine learning derived from a philosophical principle known as \textit{Occam's razor} \cite{AIXI}. Although the recent scaling paradigm of AI, where performance improvements come primarily from larger models and more compute, appears to go against Occam's razor, the principle continues to be valid and is still assumed to be important. It creates a paradox leading to arguably the most important unanswered question in the field of machine learning, \textit{why is it that more complex models seem to produce better performance in practice}? More precisely, the simplicity assumption here is that if $M$ and $M'$ are both models of $L$ then $P$ will assign a higher probability to the less complicated model. What exactly is meant by complicated though? Interestingly, we can use logical systems to define it. This is because when we assume a finite domain, each model, $M$, can be uniquely specified by a knowledge base $L_{M}$ defined as the collection of all ground atoms that are true in the model. However, some models can be uniquely specified with shorter knowledge bases. For example, take the model which assigns all labels to all inputs, this can be defined by $\{\forall x: Y_1(x), \forall x: Y_2(x), ... , \forall x: Y_n(x)\}$, where $n$ is the number of labels. This is much shorter than specifying $Y_i(x)$ or $\neg Y_i(x)$ individually for all $Y_i$ and $x$. It gives us a convenient way to define the Kolmogorov complexity of a model. To start, we can transform a knowledge base $L$ into a single sentence $\phi(L)$ by taking the conjunction on each sentence in $L$, define the complexity of $M$ to be $k(M)=\min_{L\in \mathcal{L}_{M}} |\phi(L)|$, where $|\phi(L)|$ is the length of $\phi(L)$ and $\mathcal{L}_M$ is the set of knowledge bases which have $M$ as their only model.\\\\
We can state Property 2 precisely by:
\begin{equation}
    P_L(M')=\frac{f(k(M'))}{\sum\limits_{M: M\vDash L} f(k(M))},
\end{equation}
where $k(M)$ is the complexity of $M$ as defined above and $f$ is a strictly decreasing function. The above equation can be used to directly measure the increase in probability of a model $M$ with additional background knowledge $L'$, where $Z_L$ is a normalizing constant.

\begin{equation*}
    \begin{split}
        \frac{P_{L\cup L'}( M)}{P_L(M)} &=\frac{\frac{f(k(M))}{Z_{L\cup L'}} }{\frac{f(k(M))}{Z_L}}=\frac{Z_L}{Z_{L\cup L'}}\\
        & =\frac{\sum\limits_{M:M\vDash L} f(k(M))}{\sum\limits_{M:M\vDash L\cup L'}f(k(M))}\\
        &=1+ \frac{\sum\limits_{M\in \mathcal{M}_L\setminus \mathcal{M}_{L'}}f(k(M))}{Z_{L\cup L'}}\\
        &=1 +\frac{\sum\limits_{M\in \mathcal{M}_L\setminus \mathcal{M}_{L'}}f(k(M))}{Z_L-\sum\limits_{M\in \mathcal{M}_L\setminus \mathcal{M}_{L'}}f(k(M))}
    \end{split}
\end{equation*}
For clarity, in the above derivation we abuse notation by denoting $Z_{(L-L')}$ as $\sum\limits_{M\in \mathcal{M}_L\setminus \mathcal{M}_{L'}}f(k(m))$, to obtain: 
\begin{equation}
     \frac{P_{L\cup L'}(M)}{P_L(M)}=1+ \frac{Z_{L-L'}}{Z_L-Z_{(L-L')}}.
\end{equation}
Thus, to maximize the increase in probability of the correct model we must maximize $Z_{L-L'}$. This is done by choosing $L'$ to disqualify as many models with low complexity as possible. There may be a trade off in doing this, either disqualify a small number of medium and high complexity models or a large number of low complexity models.\\\\
In \cite{deepcheap}, it is suggested that the reason why neural networks are able to generalize is because the types of problems that we are interested in typically have low-dimensional polynomial Hamiltonians, a general property of our universe. They argue that it is in this sense that neural networks have a bias towards low complexity. It is worth questioning whether complexity in this sense comports with the symbolic complexity we have defined above. For instance, in our examples, all $x_i$ are individual elements of the domain with no inherent relation between them, but neural networks are usually trained on vectors in $\mathbb{R}^n$ meaning that the domain has additional metric structure. If $x_i$ and $x_k$ are close in Euclidean distance space then the network is more likely to assign them to the same label without the training data distinguishing them. This could be the case despite the fact that there is a symbolically less complicated solution that assigns them to different labels. Dually, we are seeing that in large language models there are far too many exceptions (close by data points with different labels) that are making it very difficult and costly to align trained neural models with sound reasoning requirements \cite{gsm-symbolic}. With all of this is mind, although Property 1 and Property 2 may not hold outright, they are expected to indicate the direction of travel for the research efforts to improving the learning and reasoning capabilities of neural networks.\\\\ 
%After all, if neural networks were more likely to learn the model with lower symbolic complexity, then softmax units would be redundant, as a model with exclusive labels is less complex than one without.
Properties 1 and 2 also exemplify the benefit of formulating semantic encoding explicitly. While this research is still in its preliminary stages, it can be seen already that one can derive properties of learning algorithms that are useful for neurosymbolic encoding. Finding general properties that would apply to any network encoding given a logic is the ultimate goal of a theory of semantic encoding. When these properties are discovered, learning algorithms can be designed to satisfy the properties leading to encoding methods with greater impact on the ability of a neural network to generalize beyond the data distribution.
\section{Conclusion}
In this paper we have examined the relationship between neural networks and logical semantics. We have done this primarily through the lens of neural encodings but have expanded the scope to give semantic descriptions of deep learning in general. We have described the components of semantic encoding and discussed the philosophical and technical issues that need to be addressed to make use of semantic encoding in practice, reconciling learning and reasoning. There is much fertile ground here for new research as a dedicated theory has the potential to not only inform neural encoding techniques but deep learning in general. This is still an underexplored area that with the development of general soft-learning techniques \cite{deeplog} we can begin to answer key questions about the usefulness of knowledge consolidation as part of the learning process. Future work could focus on the development of a set of general encoding principles allowing any logic to be encoded in a neural network in ways that preserve important information about the data. A recent investigation of the principle of \textit{fibring} both in logic and neural networks and their combination has showed interesting new relationships between transformers and graph neural networks with modal logic \cite{harzli2025neuralnetworkslogicaltheories}. The main take away is that the semantics of a logical system can be seen as a point of interface between neural networks and logic, as neural networks attempt to learn information about the real world from a dataset, information which can be described using logic, the calculus of computer science. The combination of logic and neural networks in neurosymbolic AI systems has the hallmarks of being more powerful, flexible, and generalizable than either of the two approaches separately.
\bibliographystyle{alpha}
\bibliography{sample}

\end{document}